\documentclass[10pt,journal,compsoc]{IEEEtran}

\ifCLASSOPTIONcompsoc
  \usepackage[nocompress]{cite}
\else
  \usepackage{cite}
\fi
\usepackage[printwatermark]{xwatermark}
\usepackage{xcolor}
\usepackage{graphicx}
\usepackage{lipsum}

\ifCLASSINFOpdf
 \usepackage{algpseudocode}
\usepackage{algorithm}
\usepackage{mathtools}
\usepackage{amsmath}
\usepackage{amssymb}
\usepackage{xcolor}
\newcommand{\sign}{\text{sign}}
\usepackage{amsthm}
  \newtheoremstyle{dotless}{}{}{\itshape}{}{\bfseries}{}{ }{}
  \theoremstyle{dotless}
  
  \newtheorem{prop}{Proposition}

\else
\fi
\usepackage{bm}
\usepackage{amsmath}
\newcommand\norm[1]{\left\lVert#1\right\rVert}
\begin{document}
%
\title{$\ell_p$-Norm Multiple Kernel One-Class Fisher Null-Space}
%
%
%
%

\author{Shervin~Rahimzadeh~Arashloo~\IEEEmembership{}
\IEEEcompsocitemizethanks{\IEEEcompsocthanksitem S.R. Arashloo is with the Centre for Vision, Speech and Signal Processing (CVSSP), University of Surrey, Guildford, Surrey, GU2 7XH, U.K.\protect\\
E-mail: s.rahimzadeh@surrey.ac.uk.}
\thanks{This work has been submitted to the IEEE for possible publication. Copyright may be transferred without notice, after which this version may no longer be accessible.}}

%
%

\markboth{Journal of \LaTeX\ Class Files,~Vol.~14, No.~8, August~2015}%
{Shell \MakeLowercase{\textit{et al.}}: Bare Advanced Demo of IEEEtran.cls for IEEE Computer Society Journals}
%



\IEEEtitleabstractindextext{%
\begin{abstract}
The paper addresses the multiple kernel learning (MKL) problem for one-class classification (OCC). For this purpose, based on the Fisher null-space one-class classification principle, we present a multiple kernel learning algorithm where a general $\ell_p$-norm constraint ($p\geq1$) on kernel weights is considered. We cast the proposed one-class MKL task as a min-max saddle point Lagrangian optimisation problem and propose an efficient method to solve it. An extension of the proposed one-class MKL approach is also considered where several related one-class MKL tasks are learned jointly by constraining them to share common kernel weights. 

An extensive assessment of the proposed method on a range of data sets from different application domains confirms its merits against the baseline and several other algorithms.
\end{abstract}

}

\maketitle

\IEEEdisplaynontitleabstractindextext

%
\IEEEpeerreviewmaketitle

\ifCLASSOPTIONcompsoc
\IEEEraisesectionheading{\section{Introduction}\label{sec:introduction}}
\else
\section{Introduction}
\label{sec:introduction}
\fi

%
%
%
%
\IEEEPARstart{O}{ne}-class classification (OCC) corresponds to recognition of patterns that meet a particular condition identified as normal, and discerning them from any irregular observation diverging from normality, known as anomalies, abnormalities, novelties, etc. \cite{Chandola:2009:ADS:1541880.1541882,khan_madden_2014}. Unlike the common multi-class classification paradigm, OCC mainly rests on observations from a single (typically normal/target) class for model construction. A one-class design scheme may be favoured in practice due to different reasons such as imbalanced data \cite{5128907} or difficulty in accessing representative training samples from the non-target class. The limitations in obtaining prototypical training observations may arise, for instance, because of the high cost associated with collecting samples from the non-target class, the unpredictable form of test observations, the openness of a recognition problem, etc. Such situations may arise in a variety of applications including biometric presentation attack detection \cite{7984788}, health care \cite{6566012}, surveillance \cite{4668357}, intrusion detection \cite{6846360}, safety-critical systems \cite{4694106}, fraud detection \cite{Kamaruddin:2016:CCF:2980258.2980319}, insurance \cite{7435726}, etc. As such, OCC serves as an indispensable ingredient to a diverse range of real-world systems.

There exist different studies on OCC, resulting in a plethora of different methods, including the density methods, the reconstruction-based techniques, and the boundary approaches \cite{30e27ea86}. A popular kernel-based one-class classification method is the support vector data description (SVDD) approach \cite{Tax2004,Jiang_Wang_Hu_Kakde_Chaudhuri_2019} which draws on the SVM classifier and estimates a spherical boundary to enclose positive training samples which is then used to classify test observations. A closely related method to that of SVDD is the one-class SVM (OCSVM) approach \cite{Scholkopf:2001:ESH:1119748.1119749} where the positive instances are split from the origin using a maximum margin hyperplane. Some other instances of successful kernel-based OCC methods are the GP (Gaussian Process) method \cite{KEMMLER20133507} and the one-class KPCA (Kernel Principal Component Analysis) approach \cite{HOFFMANN2007863}. Among others, the work in \cite{8099922,6619277} present a unique approach called the \textit{one-class Fisher null-space} method. In this method, the one-class classifier operates on an adaptation of the Fisher classification principle to the one-class setting, yielding a theoretically optimal Fisher criterion value. While conventionally the Fisher classifier requires both positive and negative samples for training, in the one-class Fisher null-space approach, the requirement for the existence of negative training observations is circumvented by assuming the origin as a hypothetical negative sample and training the classifier using only positive/target observations. Through a formulation in the reproducing kernel Hilbert space and by using non-linear kernel functions, the one-class Fisher null-space method exhibits flexibility to deal with inherently non-linear data. The promising performance of this approach as compared with several other alternatives in different applications \cite{DBLP:journals/corr/abs-1807-01085,6619277,8099922,8546301} has led to further developments and improvements of this method \cite{9055448}.

The formulation of the one-class Fisher null-space approach in the kernel space enables decoupling of learning the classifier from data representations while opening the possibility to benefit from powerful machine learning techniques. Nevertheless, similar to other kernel-based methods, selecting an appropriate kernel function is an important design consideration in the Fisher null-space OCC method as the kernel characterises the embedding of the observations in the feature space, which directly impacts on the performance. While, preferably, the embedding should be learnt directly from the data, in kernel-based methods, this challenging problem is usually relaxed as searching for an optimal combination of multiple kernels, each capturing a distinct view of the problem. Accessibility of multiple kernel candidates for a specific problem may arise, for example, due to multiple data representations, heterogeneous data sources, or as a result of using different kernel functions. The availability of multiple kernels and the desire to benefit from multiple information sources for an improved performance motivates the studies concerning an automatic combination of several kernels, referred to as multiple kernel learning (MKL) \cite{JMLR:v12:gonen11a,6866224,7792117,6654166,6579606,5601738,8627941}. Very often, the composite kernel is expressed in terms of a linear combination of multiple base kernels, rendering the MKL problem as one of searching for the optimal combination weights under the constraint that a valid combined kernel is derived.

Among others, the choice of a particular regularisation constraint on kernel weights is an important design issue that directly affects the performance of an MKL algorithm. Historically, an $\ell_1$-norm regularisation that leads to sparse coefficients has been one of the popular choices \cite{4912215}. Nevertheless, sparse MKL models have been frequently observed to be outperformed by their non-sparse counterparts as they tend to discard some useful representations from the model \cite{JMLR:v13:yan12a,JMLR:v12:kloft11a,35392}. In multi-class scenarios and in response to the quest for a model with the potential of improving the accuracy over a uniform-weight kernel combination rule, an $\ell_p$- or a mixed-norm regularisation may be considered \cite{JMLR:v13:yan12a,JMLR:v12:kloft11a,8259375}. While in the multi-class case alternative MKL formulations based on the SVM or the Fisher discriminant analysis exist \cite{JMLR:v12:gonen11a,JMLR:v13:yan12a,JMLR:v12:kloft11a,Lanckriet2004}, for OCC, the MKL methods appear to be relatively scarce and formulated in terms of one-class SVM's with no earlier study addressing the one-class MKL problem within the \textit{Fisher} framework. Moreover, in the multi-class classification paradigm, different regularisation constraints including the $\ell_p$- and the mixed-norm have been considered whereas in the one-class setting, the focus has been primarily on an $\ell_1$-norm regularisation \cite{rak08,loosli:hal-01593595,3994}.

In this work, we consider the one-class MKL problem and address it in the context of the kernel Fisher null-space method \cite{8099922,6619277}, and in particular, in terms of its regression-based reformulation \cite{9055448} for one-class classification. We formulate the one-class MKL problem by considering a general $\ell_p$-norm constraint ($p \geq 1$) on kernel weights, providing the model with the flexibility to cope with problems with different sparsity characteristics. More specifically, the inclusion of a regularisation parameter in the proposed approach allows the method to tune into the intrinsic sparsity of a specific one-class task to improve the performance. Additionally, we illustrate that when several one-class problems are presumed to be related in terms of their kernel space representations, the proposed approach may be naturally extended to \textit{jointly} learn multiple OCC MKL problems by constraining them to share common kernel weights.
\subsection{Contributions}
The major contributions of the present work are as follows.
\begin{itemize}
\item Based on the Fisher null-space one-class classification method \cite{8099922,6619277,9055448}, we propose an $\ell_p$-norm one-class multiple kernel learning approach ("$\ell_p$ MK-FN") for one-class classification and formulate it as a saddle point Lagrangian optimisation problem.
\item We introduce an effective method to solve the min-max optimisation problem associated with the $\ell_p$ MK-FN method.
\item We extend the $\ell_p$ MK-FN method to learn several related one-class problems jointly ("Joint $\ell_p$ MK-FN") by coupling different OCC problems in terms of kernel weights.
\item We conduct a thorough evaluation of the proposed approach on a range of different databases from different application domains and provide a comparison to the baseline and several other one-class multiple kernel fusion methods, including the SVM-based MKL techniques and end-to-end one-class deep learning approaches to illustrate the merits of the proposed method.
\end{itemize}

In essence, the current study benefits from and relates to the previous studies on multiple kernel learning including those in \cite{JMLR:v12:kloft11a,3994}. Nevertheless, the specific focus on one-class classification in this work and the extension to learning multiple related one-class MKL problems jointly, along with the relevant experimental analysis and findings, including the one that multiple kernel Fisher null-space approach shows a tendency to perform better than the one-class SVM-based multiple kernel learning, as well as one-class deep and non-deep learning approaches for video/image novelty, abnormality, and attack detection, is significant, considering that SVM-based and deep learning methods are widely deployed in most OCC systems. A joint learning formulation of the MKL one-class Fisher null-space approach is also important from a practical point of view as it provides the opportunity to deal with the problem of data scarceness in OCC tasks while making it possible to completely avoid the learning stage for a one-class task given that a joint multiple kernel learning of several related and similar problems has been already performed.
\subsection{Organisation}
The remainder of the article is structured as detailed next. In Section \ref{RW}, a review of the relevant work on one-class multiple kernel learning is provided. In Section \ref{BG}, we first provide a brief background on the kernel Fisher null-space approach \cite{8099922,6619277} for one-class classification followed by a description of its regression-based reformulation for OCC as considered in \cite{9055448}. In Section \ref{lpmkl}, we present our new $\ell_p$-norm multiple kernel Fisher null-space ($\ell_p$ MK-FN) approach for one-class classification. In Section \ref{mlpmkl}, an extension of the $\ell_p$ MK-FN method for \textit{joint} multiple kernel learning of several related one-class problems is introduced. The results of an extensive experimental evaluation of the proposed method along with a comparison against other techniques on different datasets is presented in Section \ref{EE}. Finally, in Section \ref{C}, conclusions are drawn.
\section{Related Work}
\label{RW}
As the focus of this study is on one-class classification, we skip a review of the multi-class multiple kernel learning methods. One may consult \cite{JMLR:v12:gonen11a,JMLR:v13:yan12a,6654166} for a taxonomy and comparison of multi-class multiple kernel learning algorithms.

For one-class classification, there exist different MKL algorithms. As an instance, in \cite{rak08}, the MKL problem is addressed through through a weighted 2-norm regularisation constraint on kernel weights that promotes sparse weights. The authors proposed an algorithm called SimpleMKL to solve the formulated MKL problem. Although the method is mainly developed for binary classification based on an SVM formulation, the extension to the one-class setting for an $\ell_1$-norm OCSVM \cite{Scholkopf:2001:ESH:1119748.1119749} is also discussed. Other work \cite{loosli:hal-01593595} presents a modification of the SVDD approach to multiple kernels and applies the methodology presented in the SimpleMKL approach to enforce an $\ell_1$-norm regularisation on kernel weights. By observing that in an SVM formulation a larger number of support vectors lead to tighter solution boundaries, the study also considers the so-called \textit{slim} variants of the multiple kernel SVDD and OCSVM methods by modifying the objective function in a way that tighter solutions are favoured over loose ones. For comparison purposes, an equivalent method is also developed for the One-Class SVM and compared for 3D shape analysis and detecting outliers where the proposed approaches are deployed for both supervised and unsupervised outlier detection. The work in \cite{3994} focuses on semi-infinite linear programming formulations of the $\ell_1$-norm SVM-based multi-class MKL problem. The proposed formulation is shown to be solved efficiently via recycling the standard support vector machine implementations. Possible generalisations of this approach to handle a wider set of problems including regression and one-class classification are discussed. Nevertheless, no experimental analysis for one-class classification is provided in this work. In a different study \cite{GONEN2013795}, and in contrary to the previous studies that assign a common weight to the entire observation space, instead of assigning a common weight spanning the whole observation space, a localized multiple kernel learning algorithm, comprised of a learning method and a parametrised function for assigning local weights to the kernels is presented. The two ingredients of the algorithm are then trained using an alternating optimisation approach in a coupled fashion. An $\ell_1$-norm regularisation is implicitly assumed by choosing specific gating functions for assigning local weights. The work in \cite{JMLR:v12:kloft11a} studies the MKL problem with a special focus on an SVM-based formulation and provide insights and connections between different MKL formulations and propose two different optimisation strategies for the MKL problem subject to an $\ell_p$-norm regularisation constraint. Although the main focus of the work in \cite{JMLR:v12:kloft11a} is on two-/multi-class classification, one-class classification is also discussed as a special case. Yet, no experimental analysis for the one-class scenario is carried out.

Despite that fact that in some of the studies mentioned above generic loss functions are discussed \cite{JMLR:v12:kloft11a}, they primarily concentrate on the hinge loss function. In this regard, the corresponding multiple kernel learning algorithms principally correspond to multiple kernel support vector machines. In comparison to support vector machines where the soft margin is maximised, in the Fisher discriminant analysis the ratio of the between- and the within-class scatter in a subspace is maximised. The outstanding classification performance of the Fisher classification principle as compared with other competitors, including the support vector machines has established the Fisher classifier a favourable choice for many classification problems over multiple decades. Its extension to the null-space variant which attains the theoretically optimal value of the Fisher ratio has made it an even more appealing classification approach not only for multi-class problems but also for one-class classification tasks by providing comparable or even superior performance compared with not only SVM but also deep end-to-end learning methods when operating on pre-trained deep representations.

Considering the body of existing work on MKL for OCC, first, one notices that the existing one-class classification MKL methods focus on SVM-based formulations. In contrast, in this study, a Fisher classification principle for multiple kernel learning in a one-class setting is considered. Moreover, we consider a general $\ell_p$-norm constraint on kernel weights in contrast to the majority of other studies focusing on a fixed-norm (typically an $\ell_1$-norm) regularisation in the one-class setting. And last but not the least, we show that the proposed method can be naturally extended to jointly learn several related OCC problems and discuss its advantages. In this respect, there exists no earlier study on \textit{joint} one-class multiple kernel learning for several related one-class problems. As will be discussed further in the subsequent sections, motivated by the success of the one-class Fisher null approach for OCC, the current study explores the $\ell_p$-norm MKL paradigm for the one-class Fisher null classification principle and presents an MKL algorithm for this purpose. In this formalism, the current study illustrates that an optimal multiple kernel combination of pre-trained features obtained from the common CNNs currently in use, yields a very good performance for OCC while it is also applicable to non-deep features.
\section{Preliminaries}
\label{BG}
In this section, first, an overview of the one-class Fisher null-space approach \cite{8099922,6619277} is presented. Then, we review how the Fisher null-space method is posed as a regression problem in our earlier study \cite{9055448} for one-class classification.
\subsection{One-Class Fisher Null-Space}
The one-class kernel Fisher null-space approach for one-class classification \cite{8099922,6619277} is based on the Fisher classification principle. In a Fisher classifier, one seeks the maximiser of the objective function $\mathbf{G(\boldsymbol\upsilon)}$:
\begin{eqnarray}
 \operatorname*{arg\,max}_{\boldsymbol\upsilon}\mathbf{G(\boldsymbol\upsilon)}=\operatorname*{arg\,max}_{\boldsymbol\upsilon}\frac{\mathbf{\boldsymbol\upsilon^\top S_b \boldsymbol\upsilon}}{\mathbf{\boldsymbol\upsilon^\top S_w \boldsymbol\upsilon}}
\label{null}
\end{eqnarray}
\noindent $\mathbf{S_w}$ and $\mathbf{S_b}$ stand for the within-class and the between-class scatter matrices, respectively, while $\mathbf{\boldsymbol\upsilon}$ represents one axis of the subspace. The within-class and the between-class scatter matrices are defined as
\begin{eqnarray}
\nonumber \mathbf{S_b} &=& (\mathbf{m}_2-\mathbf{m}_1)(\mathbf{m}_2-\mathbf{m}_1)^\top\\
\nonumber \mathbf{S_w}&=&\sum_{c={1,2}}\sum_{\mathbf{x}\in c}(\mathbf{x}-\mathbf{m}_c)(\mathbf{x}-\mathbf{m}_c)^\top\\
\end{eqnarray}
where index $c$ runs over all classes while $\mathbf{m}_c$ stands for the mean of class $c$. The sole maximiser of the problem in Eq. \ref{null} is given as the leading eigenvector of the generalised eigenvalue problem
\begin{eqnarray}
\mathbf{S_b\boldsymbol\upsilon}=\lambda \mathbf{S_w\boldsymbol\upsilon}
\label{geig}
\end{eqnarray}
After deriving $\boldsymbol\upsilon$, a sample $\mathbf{x}$ is projected onto the Fisher subspace as
\begin{eqnarray}
\mathbf{y} = \mathbf{\boldsymbol\upsilon}^\top \mathbf{x}
\label{nullproj}
\end{eqnarray}
The theoretically \textit{optimal} discriminant that maximises the criterion function $\mathbf{G}(\boldsymbol\upsilon)$ is the one that leads to a between-class scatter which is positive and a within-class scatter of zero, i.e.:
\begin{eqnarray}
\mathbf{\boldsymbol\upsilon^\top S_b \boldsymbol\upsilon} > 0\\
\nonumber \mathbf{\boldsymbol\upsilon^\top S_w \boldsymbol\upsilon} = 0
\label{nullF}
\end{eqnarray}
The one-class Fisher null-space method \cite{8099922,6619277} is an adaptation of the Fisher criterion to the one-class setting that solves the Fisher criterion to optimality. In a one-class setting where only positive training instances are available, in the one-class Fisher null-space method, similar to some other studies \cite{Scholkopf:2001:ESH:1119748.1119749}, the negative class is represented by a single hypothetical example lying at the origin. The goal is then to find a mapping such that all target/normal training instances are projected onto the same point distinct from the origin. Such a projection is formulated in a reproducing kernel Hilbert space (RKHS) using non-linear kernel functions to achieve the flexibility to handle data with inherently non-linear nature. The remarkable performance of this method for one-class classification as compared with several other alternatives is verified in different studies \cite{9055448,6619277,8099922,8546301,DBLP:journals/corr/abs-1807-01085}.
\subsection{Regression-Based One-Class Fisher Null-Space}
Finding the optimal projection corresponding to the one-class Fisher null-space method incorporates eigen-decomposition of dense matrices \cite{8099922,6619277}. As the corresponding eigen-decompositions are computationally demanding, a reformulation of this method in terms of regression in the RKHS is considered in \cite{9055448}. As discussed in \cite{9055448}, a reformulation of the Fisher null-space method in terms of regression for OCC not only avoids the computationally demanding eigen-decomposition of dense matrices but also paves the way to impose different regularisations on the solution to improve the generalisation performance. A brief overview of the reformulation of the kernel Fisher null-space method in terms of regression for one-class classification, as discussed in \cite{9055448}, is provided next.

Let us assume that there exist $n$ target (positive) training observations $x_i$'s, $i=1,\dots,n$. As noted earlier, the goal in the one-class kernel Fisher null-space approach is to map all the target training observations $x_i$'s onto the same point in the subspace, distinct from the origin. Without loss of generality, let us assume that all samples are to be mapped onto point $1$. To realise such a projection, instead of following an eigen-decomposition approach as practiced in \cite{8099922,6619277}, the following proposition  \cite{9055448} may be used:
\begin{prop}
Let $\boldsymbol{\phi}(x_i)$ denote the feature vector associated with $x_i$ in the RKHS and $d>n$ (where d is the RKHS dimensionality). If $\mathbf{w}$ is the optimal solution to the problem
\begin{eqnarray}
\min_\mathbf{w} \frac{1}{n}\sum_{i=1}^n(1-\boldsymbol\phi(\mathbf{x}_i)^\top\mathbf{w})^2
\label{unreg}
\end{eqnarray}
then the projection $\boldsymbol\phi(.)^\top\mathbf{w}$ maps all target training samples onto point 1, i.e. $\boldsymbol\phi(\mathbf{x}_i)^\top\mathbf{w}=1, i=1,\dots,n$.
\end{prop}
\noindent A proof is provided in Section I of the supplementary material.\\
Since through projection $\boldsymbol\phi(.)^\top\mathbf{w}$ all positive training samples are mapped onto the same point (i.e. $1$), the within-class variance for the positive class shall be zero. Moreover, the within-class variance of the negative training set is also zero since the negative class is represented using a single artificial observation lying at the origin. Hence, the total within-class scatter will be zero. Furthermore, as the projection of the positive training instances (i.e. $1$) is distinct from that of the negative training set (i.e. the origin), the projection above provides a between-class scatter which is positive. Consequently, it corresponds to a Fisher null-space projection for one-class classification (further details are provided in Section II of the supplementary material).

In \cite{9055448}, a regularisation scheme is employed to constrain the solution of Eq. \ref{unreg}. In essence, regularising the solution enforces some limitation on the functional space via incorporating a penalty term so that some regions of the solution space are less favoured. In particular, in \cite{9055448} a Tikhonov regularisation is deployed to penalise coefficients with larger magnitude to derive a more parsimonious solution. The impact of a Tikhonov regularisation is to obtain a smoother projection function \cite{10.1007/978-3-540-87536-9_23} with better generalisation capabilities \cite{9055448}. A Tikhonov regularisation on the solution of Eq. \ref{unreg} is applied as
\begin{eqnarray}
\min_\mathbf{w} \norm{\mathbf{w}}_2^2+\frac{\theta}{n}\sum_{i=1}^n(1-\boldsymbol\phi(\mathbf{x}_i)^\top\mathbf{w})^2
\label{OCKR}
\end{eqnarray}
\noindent where $\theta$ is a regularisation parameter. The projection $\boldsymbol\phi(.)^\top\mathbf{w}$ where $\mathbf{w}$ minimses the problem above corresponds to the Tikhonov-regularised one-class kernel Fisher null-space projection for OCC and is shown to outperform several other one-class classification methods in different scenarios \cite{9055448}. In kernel-based methods, it is common practice and typically more convenient to work in the dual space, as discussed in the next section.
\section{$\ell_p$-Norm Multiple Kernel One-Class Fisher Null-Space}
\label{lpmkl}
The one-class multiple kernel learning algorithm proposed in this work is based on the dual form of the regularised kernel Fisher null-space method. For deriving the dual of the optimisation problem in Eq. \ref{OCKR}, let $\delta = n/\theta$. The unconstrained optimisation problem in Eq. \ref{OCKR} may then be written as a constrained optimisation problem as
\begin{eqnarray}
\nonumber & \min_{\mathbf{w}} \mathbf{w}^\top\mathbf{w}+\frac{1}{\delta}\|\boldsymbol\zeta\|_2^2\\
&\text{s.t. } \boldsymbol\zeta = \mathbf{1} - \Phi(\mathbf{X})^\top\mathbf{w}
\end{eqnarray}
\noindent where $\Phi(\mathbf{X})$ stands for the matrix of training samples in the RKHS and $\mathbf{1}$ is an $n$-vector of ones. The Lagrangian for the constrained optimisation problem above may be formed as
\begin{eqnarray}
\mathcal{L} = \mathbf{w}^\top\mathbf{w}+\frac{1}{\delta}\|\boldsymbol\zeta\|_2^2+\boldsymbol\rho^\top(\mathbf{1} - \Phi(\mathbf{X})^\top\mathbf{w} - \boldsymbol\zeta)
\label{lag1}
\end{eqnarray}
where $\boldsymbol\rho$ is the Lagrange multiplier. In order to find the optimal $\mathbf{w}$, one needs to solve the dual problem. For this purpose, one first needs to minimise Eq. \ref{lag1} in $\mathbf{w}$ and $\boldsymbol\zeta$ and then maximize it in $\boldsymbol\rho$. In order to minimise $\mathcal{L}$ w.r.t. $\mathbf{w}$ and $\boldsymbol\zeta$, the corresponding partial derivatives are set to zero:
\begin{eqnarray}
\frac{\partial \mathcal{L}}{\partial \mathbf{w}} = 2\mathbf{w}-\Phi(\mathbf{X})\boldsymbol\rho = 0
\end{eqnarray}
and hence
\begin{eqnarray}
\mathbf{w} = \frac{1}{2}\Phi(\mathbf{X})\boldsymbol\rho
\end{eqnarray}
Plugging $\mathbf{w}$ into Eq. \ref{lag1} yields
\begin{eqnarray}
\mathcal{L} = \frac{-1}{4}\boldsymbol\rho^\top\Phi(\mathbf{X})^\top\Phi(\mathbf{X})\boldsymbol\rho+1/\delta\boldsymbol\zeta^\top\boldsymbol\zeta+\boldsymbol\rho^\top\mathbf{1}-\boldsymbol\rho^\top\boldsymbol\zeta
\label{lag11}
\end{eqnarray}
Taking the partial derivative of $\mathcal{L}$ w.r.t. $\boldsymbol\zeta$ and setting it to zero yields
\begin{eqnarray}
\boldsymbol\zeta = \frac{\delta}{2}\boldsymbol\rho
\end{eqnarray}
By plugging $\boldsymbol\zeta$ into Eq. \ref{lag11} one obtains
\begin{eqnarray}
\nonumber \mathcal{L} = \frac{-1}{4}\boldsymbol\rho^\top\mathbf{K}\boldsymbol\rho-\frac{\delta}{4}\boldsymbol\rho^\top\boldsymbol\rho+\boldsymbol\rho^\top\mathbf{1}
\end{eqnarray}
\noindent where in the last equation the definition of a kernel matrix ($\mathbf{K}$) is used. A change of variables as $\boldsymbol\alpha = \frac{1}{2}\boldsymbol\rho$ and maximising over the dual variable gives the dual problem as
\begin{eqnarray}
\max_{\boldsymbol{\alpha}}-\boldsymbol{\alpha}^\top\mathbf{K}\boldsymbol{\alpha}-\delta \boldsymbol{\alpha}^\top \boldsymbol{\alpha}+2\boldsymbol{\alpha}^\top\mathbf{1}
\label{dau}
\end{eqnarray}
In this work, the composite kernel is presumed to be a linear combination of non-negatively weighted kernels, subject to an $\ell_p$-norm constraint on kernel weights. Assuming $J$ base kernels to be combined, the kernel matrix $\mathbf{K}$ in Eq. \ref{dau} will be replaced by $\sum_{j=1}^J\beta_j\mathbf{K}_j$ where $\beta_j$'s denote the non-negative kernel weights. Replacing $\mathbf{K}$ by $\sum_{j=1}^J\beta_j\mathbf{K}_j$ and optimising over $\boldsymbol\beta$ (a vector collection of kernel weights) subject to the non-negativity and $p$-norm constraints, the optimisation problem for the proposed $\ell_p$-norm multiple kernel one-class Fisher null-space method shall be derived as a saddle point problem:
\begin{eqnarray}
\nonumber \min_{\boldsymbol{\beta}}\max_{\boldsymbol{\alpha}}&-\boldsymbol{\alpha}^\top(\sum_{j=1}^J\beta_j\mathbf{K}_j)\boldsymbol{\alpha}-\delta \boldsymbol{\alpha}^\top \boldsymbol{\alpha}+2\boldsymbol{\alpha}^\top\mathbf{1}\\
\nonumber &\text{s.t. }
\nonumber \boldsymbol\beta \geq 0, \norm{\boldsymbol{\beta}}^p_p \leq 1\\
\label{saddle1}
\end{eqnarray}
\noindent where $\boldsymbol\beta \geq 0$ is meant element-wise. Under the condition that each individual kernel matrix $\mathbf{K}_j$ is positive semidefinite, the non-negativity constraint $\boldsymbol\beta \geq 0$ guarantees that the combined kernel matrix is also positive semidefinite and is thus a valid kernel matrix while $\norm{\boldsymbol{\beta}}^p_p\leq 1$ imposes a general $\ell_p$-norm constraint on kernel weights. As noted previously, the introduction of a $p$-norm regularisation on kernel weights is beneficial as it allows the model to adjust to the underlying sparsity structure of the problem under consideration. This is in contrast to a fixed-norm regularisation which may not be able to effectively represent the intrinsic sparsity of the problem. Similar observations have been made in the multi-class scenario where a variable-norm regularisation has been found to be advantageous compared to fixed-norm solutions \cite{JMLR:v12:kloft11a,JMLR:v13:yan12a}.
\section{Optimisation}
\begin{figure}[!t]
\centering
\includegraphics[width=.9in]{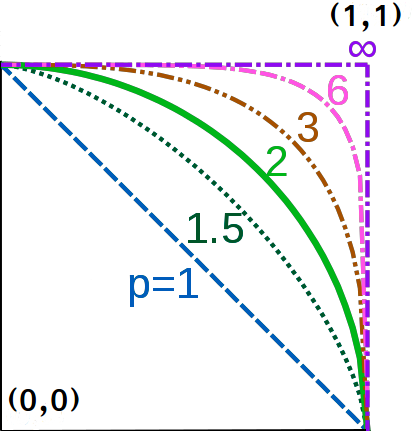}
\caption{The unit $\ell_p$-norm balls (in 2D) for $p\geq 1$ in the first quadrant.}
\label{lpspace}
\end{figure}

Let $\mathbf{u}$ be a $J$-element vector the $j^{th}$ element of which is defined as $u_j = \boldsymbol\alpha^\top\mathbf{K}_j\boldsymbol\alpha$. The optimisation problem in Eq. \ref{saddle1} may now be expressed as
\begin{eqnarray}
\nonumber \min_{\boldsymbol{\beta}}\max_{\boldsymbol{\alpha}}-\delta \boldsymbol{\alpha}^\top \boldsymbol{\alpha}+2\boldsymbol{\alpha}^\top\mathbf{1}-\boldsymbol{\beta}^\top\mathbf{u}\\
\nonumber\text{s.t. }
\nonumber \boldsymbol\beta \geq 0, \norm{\boldsymbol{\beta}}^p_p\leq 1\\
\label{beta1}
\end{eqnarray}
For fixed $\boldsymbol\alpha$, the objective function is linear in $\boldsymbol\beta$ and the constraint set for $p\geq 1$ (the focus of this study) forms a closed convex set, Fig. \ref{lpspace}. As a result, for fixed $\boldsymbol\alpha$, the cost function is convex in $\boldsymbol\beta$. On the other hand, for fixed $\boldsymbol\beta$, the objective function is concave in $\boldsymbol\alpha$. As a result, according to the generalised minimax theorem \cite{DPa95,VonNeumann1971}, Eq. \ref{beta1} may be equivalently written as
\begin{eqnarray}
\nonumber \max_{\boldsymbol{\alpha}}-\delta \boldsymbol{\alpha}^\top \boldsymbol{\alpha}+2\boldsymbol{\alpha}^\top\mathbf{1}+ \min_{\boldsymbol{\beta}} -\boldsymbol{\beta}^\top\mathbf{u}&\\
\nonumber\text{s.t. }
\nonumber \boldsymbol\beta \geq 0, \norm{\boldsymbol{\beta}}^p_p& \leq 1\\
\label{saddle12}
\end{eqnarray}
That is, one may exchange the order of optimisation w.r.t. $\boldsymbol\beta$ and $\boldsymbol\alpha$. In order to perform optimisation in $\boldsymbol\beta$, the Lagrangian of the minimisation problem may be formed as
\begin{eqnarray}
\mathcal{L} = -\boldsymbol\beta^\top(\mathbf{u}+\boldsymbol{\mu})+\gamma(\norm{\boldsymbol\beta}_p^p-1)
\end{eqnarray}
where $\gamma\geq0$ and $\boldsymbol\mu\geq0$ are Lagrange multipliers. It may be easily verified that the Slater's condition holds and the KKT conditions are
\begin{flalign}
\nonumber &\nabla_{\boldsymbol\beta} \mathcal{L} = -(\mathbf{u}+\boldsymbol\mu)+\gamma p |\boldsymbol\beta|^{p-1}\odot\sign(\boldsymbol\beta)=0 \\
\nonumber &\boldsymbol\mu^\top\boldsymbol\beta = 0 \\
\nonumber &\boldsymbol\beta\geq 0 \\
\nonumber &\gamma(\norm{\boldsymbol\beta}_p^p-1)=0\\
\end{flalign}
\noindent where $\odot$ denotes the Hadamard (element-wise) multiplication. From the first KKT condition one obtains
\begin{eqnarray}
|\boldsymbol\beta|^{p-1}\odot\sign(\boldsymbol\beta)=\frac{\mathbf{u}+\boldsymbol\mu}{\gamma p}
\label{betaf}
\end{eqnarray}
Note that if $\gamma=0$, then the $p$-norm constraint would not be satisfied and the optimisation problem would be unbounded. Consequently, $\gamma>0$, and thus, the equation above is well defined. At the optimum $\boldsymbol\beta$ is non-negative ($\boldsymbol\beta\geq 0$) and hence
\begin{eqnarray}
\boldsymbol\beta={\Big(\frac{\mathbf{u}+\boldsymbol\mu}{\gamma p}\Big)}^{1/(p-1)}
\label{betaraw}
\end{eqnarray}
By plugging this into the second KTT condition one obtains
\begin{eqnarray}
\boldsymbol\mu^\top{\big[\frac{\mathbf{u}+\boldsymbol\mu}{\gamma p}\big]}^{1/(p-1)}=0
\end{eqnarray}
Next, we will show that for $\boldsymbol\mu^\top{\big[\frac{\mathbf{u}+\boldsymbol\mu}{\gamma p}\big]}^{1/(p-1)}=0$ to hold, one must have $\boldsymbol\mu=0$. For the proof, we use contradiction. Let us assume that not all elements of $\boldsymbol\mu$ are zero and $\boldsymbol\mu$ includes one strictly positive element. Under this assumption, since $\gamma$, $p$ and $\mathbf{u}$ are non-negative, the corresponding element in ${\big[\frac{\mathbf{u}+\boldsymbol\mu}{\gamma p}\big]}^{1/(p-1)}$ shall be strictly positive too. In this case, the inner product of $\boldsymbol\mu$ and $\boldsymbol\beta$ would be strictly positive which contradicts the initial requirement of $\boldsymbol\mu^{\top}\boldsymbol\beta=0$. As a result, $\boldsymbol\mu$ cannot include any strictly positive elements, and since it is non-negative, we conclude that $\boldsymbol\mu=0$.

Due to the form of the minimisation problem w.r.t. $\boldsymbol\beta$, it is clear that the elements of $\boldsymbol\beta$ must be chosen as large as possible at the optimum which leads to the maximisation of the $p$-norm. The maximum of the $p$-norm happens when $\gamma>0$, requiring the $p$-norm constraint to hold as equality, and thus from the last KKT condition $\norm{\boldsymbol\beta}_p=1$. Since $\gamma>0$, $\boldsymbol\mu=0$ and $\norm{\boldsymbol\beta}_p=1$, $\boldsymbol\beta$ is derived as
\begin{eqnarray}
\boldsymbol\beta = \frac{\mathbf{u}^{1/(p-1)}}{\norm{\mathbf{u}^{1/(p-1)}}_p}
\label{betaupdate1}
\end{eqnarray}
The relation above is well defined for $1<p<\infty$. The degenerate case of $p=1$ and also the case of $p\to+\infty$ are analysed next.\\
$\mathbf{p\to1^+}$:
Let us consider the numerator of the quotient in Eq. \ref{betaupdate1} in when $p\to1^+$. Denoting the maximum element of $\mathbf{u}$ as $u_{max}$ we have
\begin{flalign}
\nonumber &\lim_{p\to 1^+} \mathbf{u}^{1/(p-1)}\\
\nonumber &=\lim_{p\to 1^+}\big[u_1,\dots,u_{max},\dots,u_J\big]^{1/(p-1)}\\
&=\lim_{p\to 1^+}u_{max}^{1/(p-1)}\big[(\frac{u_1}{u_{max}})^{1/(p-1)},\dots,1,\dots,(\frac{u_J}{u_{max}})^{1/(p-1)}\big]
\end{flalign}
Note that all elements of the vector above are smaller than one except for the element at the position of the maximum element which is equal to $1$. Moreover, $\lim_{p\to 1^+}1/(p-1)\to+\infty$ and consequently, in the limit when $p\to1^+$, all the elements of the vector shrink to zero except for the maximum element. As a result, when $p\to1^+$, $\boldsymbol\beta$ shall only have a single non-zero element of $1$ at the position of the maximum element of $\mathbf{u}$.
\\
$\mathbf{p\to+\boldsymbol\infty:}$
In this case, we have $\lim_{p\to+\infty}1/(p-1)=0$. As a result, all the elements of $\mathbf{u}$ are raised to zero. Assuming that the elements of $\mathbf{u}$ are non-zero (this is the case when the kernel matrix is positive definite), $\boldsymbol\beta$ is then derived as a uniform vector with a unit $p$-norm.

Up to this point, a procedure to determine $\boldsymbol\beta$ is presented. Once $\boldsymbol\beta$ is determined, the objective function should be maximised in $\boldsymbol\alpha$. Setting the partial derivative of the cost function with respect to $\boldsymbol\alpha$ to zero yields
\begin{eqnarray}
\boldsymbol\alpha = \Big(\delta \mathbf{I}+\sum_{j=1}^J\beta_j\mathbf{K}_j\Big)^{-1}\mathbf{1}
\label{alphasingle2}
\end{eqnarray}
In the relation above, $\boldsymbol\alpha$ is given in terms of $\boldsymbol\beta$. If $\boldsymbol\beta$ was independent of $\boldsymbol\alpha$, then the optimal $\boldsymbol\alpha$ could have been directly found. However, due to Eq. \ref{betaupdate1}, $\boldsymbol\beta$ is given in terms of $\mathbf{u}$ which is a function of $\boldsymbol\alpha$. In other words, in Eq. \ref{alphasingle2}, $\boldsymbol\alpha$ is defined in terms of itself which necessitates a different approach for optimisation. In this work, a fixed-point iteration procedure \cite{NM} is followed to solve for $\boldsymbol\alpha$. More specifically, let us define $f(\boldsymbol\alpha)=\Big(\delta \mathbf{I}+\sum_{j=1}^J\beta_j\mathbf{K}_j\Big)^{-1}\mathbf{1}$, and hence at the optimum it must hold that $\boldsymbol\alpha=f(\boldsymbol\alpha)$. In order to determine $\boldsymbol\alpha$, we look for the fixed-point of function $f(\boldsymbol\alpha)$. That is, at each iteration, using the current estimate of $\boldsymbol\alpha$, $f(\boldsymbol\alpha)$ is determined which is then used as the new estimate for $\boldsymbol\alpha$. The procedure above is repeated until convergence. The approach described above is summarised in Algorithm \ref{generic} where $\boldsymbol\alpha$ is initialised using a uniform $\boldsymbol\beta$ with a unit $p$-norm. A convergence analysis of the proposed approach is provided in Section III of the supplementary material.

\begin{algorithm}[t]
\caption{$\ell_p$-Norm Multiple Kernel One-Class Fisher Null-Space}
\label{generic}
\begin{algorithmic}[1]
\State Input: kernel matrices $\mathbf{K}_j$, $j=1,\dots, J$
\State Initialisation: $\boldsymbol\alpha = \big(\delta \mathbf{I}+\sum_{j=1}^J\frac{1}{J^{1/p}}\mathbf{K}_j\big)^{-1}\mathbf{1}$
\Repeat
\State $\mathbf{u} = \big[ \boldsymbol\alpha^\top\mathbf{K}_1\boldsymbol\alpha,\dots, \boldsymbol\alpha^\top\mathbf{K}_J\boldsymbol\alpha \big]$
\State $\boldsymbol\beta = \frac{\mathbf{u}^{1/(p-1)}}{\norm{\mathbf{u}^{1/(p-1)}}_p}$
\State $\boldsymbol\alpha = \big(\delta \mathbf{I}+\sum_{j=1}^J\beta_j\mathbf{K}_j\big)^{-1}\mathbf{1}$
\Until{convergence}
\State Output: $\boldsymbol{\alpha}$ and $\boldsymbol\beta$
\end{algorithmic}
\normalsize
\end{algorithm}
\subsection{Remark} The problem in Eq. \ref{saddle1} is concave in $\boldsymbol\alpha$. This may be readily observed by plugging $\boldsymbol\beta$ given in Eq. \ref{betaupdate1} into Eq. \ref{saddle12} to obtain (see Section IV in the supplementary material for a derivation)
\begin{eqnarray}
\max_{\boldsymbol{\alpha}}-\delta \boldsymbol{\alpha}^\top \boldsymbol{\alpha}+2\boldsymbol{\alpha}^\top\mathbf{1}-\norm{\mathbf{u}}_{p/(p-1)}
\end{eqnarray}
As the elements of $\mathbf{u}$ are quadratic functions of $\boldsymbol\alpha$, and since $\mathbf{K}_j$'s, $\forall j$ are positive semidefinite matrices, the elements of $\mathbf{u}$ are convex functions of $\boldsymbol\alpha$. Moreover, as any norm ($p\geq1$) is a convex function, the problem above is concave. Consequently, one may consider a gradient ascent-type algorithm to solve for $\boldsymbol\alpha$. However, in practice, as will be demonstrated in Section \ref{RunTime}, the fixed-point iteration approach presented in Algorithm \ref{generic} is significantly faster than a gradient-based method.

Note that the method presented in \cite{JMLR:v12:kloft11a} that considers a complete dual MKL problem (not solely a dual in terms of $\boldsymbol\alpha$'s as considered in this study) may be adapted and applied to the multiple kernel one-class Fisher null-space approach. In particular, in line 5 of Algorithm \ref{generic}, the kernel mixture weight parameter ($\boldsymbol\beta$) may also be determined by adapting the analytical update formula presented in \cite{JMLR:v12:kloft11a} to the Fisher null-space framework. In this context, the formulation presented in this work is equivalent to that of \cite{JMLR:v12:kloft11a} in terms of learnt kernel mixture parameters, but can potentially lead to better numerical stability and faster convergence. In particular, in our experiments, we observed that in the one-class Fisher null-space approach, when parameter $p$ is chosen to promote a high degree of sparsity in the kernel weight vector (i.e. selected to be close to 1) the analytical update relation given in \cite{JMLR:v12:kloft11a} may be subject to numerical instability and relatively slower convergence as compared to that of this study.
\section{Extension to Joint One-Class Multiple Kernel Learning}
\label{mlpmkl}
In the discussions thus far, it was assumed that only a \textit{single} one-class problem exists. In some applications, however, there might be several \textit{related} one-class problems to be learned. In such situations, different one-class tasks may share similarities that motivate inferring common kernel weights shared across all the OCC problems. This may be useful when the training samples for some one-class problems are scarce and not representative of the entire observation space. In this case, an MKL approach when applied separately to each individual one-class task may suffer from poor generalisation on unseen test data but taking all training observations from all one-class tasks may better represent the feature space. A further appealing characteristic of a joint one-class MKL approach is that of computational efficiency. In this context, under the assumption that different one-class MKL problems are related and share common kernel weights, once a joint learning is performed for a number of OCC MKL problems, any new and similar one-class MKL problem may then benefit from the weights previously inferred, avoiding the requirement for a further MKL learning procedure. In this section, an extension of Algorithm \ref{generic} to handle the case of joint multiple kernel learning for several related OCC problems is presented.

Let us assume that there exist $C$ related one-class problems which are supposed to benefit from common kernel weights. We assume that each individual one-class problem (indexed by $c$) is specified by a distinct set of kernel matrices (i.e. $\mathbf{K}^c_j,\text{  for } j=1,\dots,J$) and a separate discriminant in the Hilbert space (i.e. $\boldsymbol{\alpha}_c$) but coupled to other OCC problems through a common kernel weight vector, i.e. $\boldsymbol{\beta}$. With these assumptions, the optimisation problem for the joint $\ell_p$-norm multiple kernel Fisher null-space may be written as
\begin{eqnarray}
\label{jointeq}
\nonumber \min_{\boldsymbol{\beta}}\max_{\boldsymbol{\alpha}_c }&\sum_{c=1}^C -\boldsymbol{\alpha}_c^\top(\sum_{j=1}^J\beta_j\mathbf{K}^c_j)\boldsymbol{\alpha}_c-\delta \boldsymbol{\alpha}_c^\top \boldsymbol{\alpha}_c+2\boldsymbol{\alpha}_c^\top\mathbf{1}_c\\
&\text{s.t. }
\boldsymbol\beta \geq 0, \norm{\boldsymbol{\beta}}^p_p\leq 1
\end{eqnarray}
\noindent where $\mathbf{1}_c$ is a vector of ones of a dimensionality suitable for the $c^{th}$ one-class problem. If the kernel weights were learned independently for each individual OCC problem, the optimisation problem above would have been decoupled into $C$ independent OCC MKL problems to each of which Algorithm \ref{generic} could be directly applied. However, the introduction of a common kernel weight vector $\boldsymbol\beta$ shared across several OCC tasks necessitates a modified optimisation procedure, discussed next.
\begin{algorithm}[t]
\caption{Joint $\ell_p$-Norm Multiple Kernel One-Class Fisher Null-Space}
\label{joint}
\begin{algorithmic}[1]
\State Input: kernel matrices $\mathbf{K}^c_j$, $j = 1,\dots, J$, $c = 1,\dots, C$
\State Initialisation:
\For{$c=1,\dots, C$}
\State $\boldsymbol\alpha_c = \Big(\delta \mathbf{I}+\sum_{j=1}^J\frac{1}{J^{1/p}}\mathbf{K}^c_j\Big)^{-1}\mathbf{1}_c$
\EndFor
\Repeat
\State $\mathbf{v} = \big[\sum_{c=1}^C\boldsymbol\alpha_c^\top\mathbf{K}^c_1\boldsymbol\alpha_c, \dots, \sum_{c=1}^C\boldsymbol\alpha_c^\top\mathbf{K}^c_J\boldsymbol\alpha_c\big]$
\State $\boldsymbol\beta = \frac{\mathbf{v}^{1/(p-1)}}{\norm{\mathbf{v}^{1/(p-1)}}_p}$
\For{$c=1,\dots, C$}
\State $\boldsymbol\alpha_c = \Big(\delta \mathbf{I}+\sum_{j=1}^J\beta_j\mathbf{K}^c_j\Big)^{-1}\mathbf{1}_c$
\EndFor
\Until{convergence}
\State Output: $\boldsymbol{\alpha}_c$'s and $\boldsymbol\beta$
\end{algorithmic}
\normalsize
\end{algorithm}
Let us define $v_j$ to be the $j^{th}$ element of the vector $\mathbf{v}$ given as $v_j = \sum_{c=1}^C\boldsymbol\alpha_c^\top\mathbf{K}^c_j\boldsymbol\alpha_c$. Applying the minimax theorem, the saddle point problem in Eq. \ref{jointeq} may be written as
\begin{eqnarray}
\nonumber \max_{\boldsymbol\alpha_c}{\sum_{c=1}^C -\delta \boldsymbol{\alpha}_c^\top \boldsymbol{\alpha}_c+2\boldsymbol{\alpha}_c^\top\mathbf{1}_c+ \min_{\boldsymbol\beta} -\boldsymbol\beta^\top\mathbf{v}}&\\
\nonumber \text{s.t.  }
\nonumber \boldsymbol\beta \geq 0, \norm{\boldsymbol{\beta}}^p_p&\leq 1\\
\label{beta2}
\end{eqnarray}
One notices the similarity of the minimisation subproblem above to that of Eq. \ref{saddle12}. Hence, following a similar procedure, $\boldsymbol\beta$ is derived as
\begin{eqnarray}
\boldsymbol\beta = \frac{\mathbf{v}^{1/(p-1)}}{\norm{\mathbf{v}^{1/(p-1)}}_p}
\label{betaupdate2}
\end{eqnarray}
In order to maximise the objective function in Eq. \ref{jointeq} in $\boldsymbol\alpha_c$'s, one may take the partial derivatives with respect to each $\boldsymbol\alpha_c$ and set them to zero to obtain
\begin{eqnarray}
\boldsymbol\alpha_c = \Big(\delta \mathbf{I}+\sum_{j=1}^J\beta_j\mathbf{K}^c_j\Big)^{-1}\mathbf{1}_c \text{,    for } c=1,\dots, C
\label{alphajoint}
\end{eqnarray}
As in the single OCC scenario, a fixed-point iteration approach may be applied to find the optimal solution. The proposed joint multiple kernel one-class Fisher null-space approach is summarised as Algorithm \ref{joint} where $\boldsymbol\alpha$ is initialised using a uniform unit $p$-norm kernel weight vector.

\section{Experimental Evaluation}
\label{EE}
In this section, the results of an experimental evaluation of the proposed approach and a comparison against other methods from the literature are provided. The kernel-based methods included in the comparison are:
\begin{itemize}
\item \textit{FN-Average}: As a first baseline method, this method corresponds to the element-wise arithmetic mean of the kernels followed by the kernel Fisher null-space approach for OCC \cite{8099922,6619277};
\item \textit{FN-Product}: As a second baseline method, this method corresponds to the element-wise geometric mean of the kernels followed by the kernel Fisher null-space approach for OCC \cite{8099922,6619277};
\item \textit{GP-Average}: corresponds to the element-wise arithmetic mean of the kernels followed by the kernel GP method for OCC \cite{KEMMLER20133507};
\item \textit{GP-Product}: corresponds to the element-wise geometric mean of the kernels followed by the kernel GP method for OCC \cite{KEMMLER20133507};
\item \textit{KPCA-Average}: corresponds to the element-wise arithmetic mean of the kernels followed by the KPCA method for OCC \cite{HOFFMANN2007863};
\item \textit{KPCA-Product}: corresponds to the element-wise geometric mean of the kernels followed by the KPCA method for OCC \cite{HOFFMANN2007863};
\item \textit{MK-SVDD}: is the multiple kernel learning method based on the SVDD approach for OCC \cite{loosli:hal-01593595};
\item \textit{Slim-MK-SVDD}: is the multiple kernel learning method based on the SVDD approach for OCC where the objective function is modified so that tight solutions are favoured over loose ones \cite{loosli:hal-01593595};
\item \textit{MK-OCSVM}: is the multiple kernel learning method based on the One-Class SVM algorithm for OCC \cite{rak08};
\item \textit{Slim-MK-OCSVM}: is the multiple kernel learning method based on the One-Class SVM algorithm for OCC where the objective function is modified so that tight solutions are favoured over loose ones \cite{loosli:hal-01593595};
\item $\ell_p$ \textit{MK-FN}: is the proposed $\ell_p$-norm multiple kernel Fisher null-space approach;
\item \textit{Joint} $\ell_p$ \textit{MK-FN}: is the proposed $\ell_p$-norm multiple kernel Fisher null-space approach where a shared kernel weight is inferred for several related one-class problems.
\end{itemize}
In addition to the aforementioned kernel-based methods, wherever applicable, we include state-of-the-art end-to-end one-class deep learning methods in the comparisons.

The remainder of this section is organised as detailed next.
\begin{itemize}
\item Section \ref{imp} provides the details of implementation.
\item In Section \ref{cb}, we analyse the convergence behaviour of the proposed approach.
\item In Section \ref{peffect}, we verify that unique choices of $p$ in the proposed $\ell_p$ MK-FN approach result in different behaviour in terms of benefiting from multiple informative kernels, or being robust against noisy kernels.
\item In Section \ref{Delft}, we evaluate the proposed $\ell_p$-norm MK-FN approach on several benchmark one-class datasets from the Delft university and compare it to the baseline as well as other one-class multiple kernel fusion methods.
\item In Section \ref{Abn}, the proposed approach is evaluated for abnormality detection and compared against the baseline as well as other one-class multiple kernel fusion and deep end-to-end OCC methods.
\item In Section \ref{ImgNovel}, the proposed approach is evaluated for one-class novelty detection and compared to the baseline as well as other one-class multiple kernel fusion and deep end-to-end OCC methods.
\item In Section \ref{Anoface}, we conduct experiments in biometric anti-spoofing (presentation attack detection) and compare the performance of the proposed approach to the baseline as well as other one-class multiple kernel fusion and deep end-to-end OCC methods.
\item Finally, in Section \ref{RunTime}, we compare the running time of different one-class MKL methods and discuss the scalability properties of the proposed approach.
\end{itemize}
\subsection{Implementation details}
\label{imp}
Motivated by earlier studies in the multi-class setting, the parameter $p$ in the proposed approach is chosen from $\{1, 32/31, 16/15, 8/7, 4/3, 2, 4, 8, 10^6\}$. The regularisation parameter $\delta$ is selected from $\{10^{-4}, 10^{-3}, 10^{-2}, 10^{-1}, 1, 10, 10^2\}\times n$ ($n$ being the number of training samples). The kernel function used in the experiments is that of a Gaussian (RBF) kernel the width of which is chosen from $\{\frac{1}{4}D, \frac{1}{2}D, D\}$ where $D$ is the average pairwise Euclidean distance between training samples. The parameters of the proposed approach are set on the validation set. The parameters of the multiple kernel learning methods based on SVM are set as suggested in \cite{loosli:hal-01593595} on the validation set.
\subsection{Convergence Behaviour}
\label{cb}
\begin{figure}[t]
\centering
\includegraphics[scale=.25]{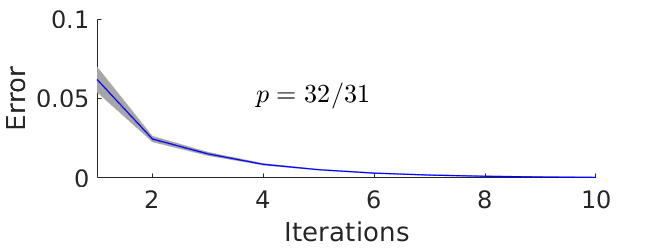}
\includegraphics[scale=.25]{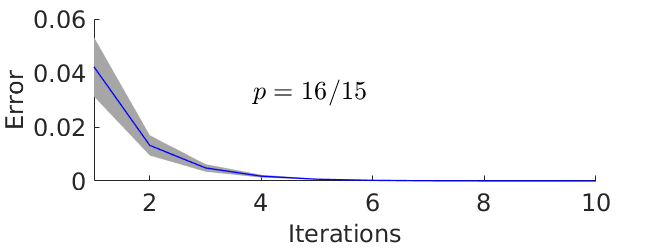}
\includegraphics[scale=.25]{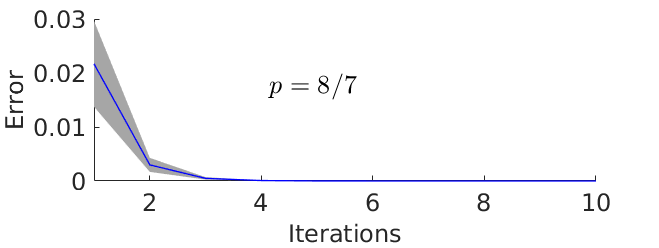}
\includegraphics[scale=.25]{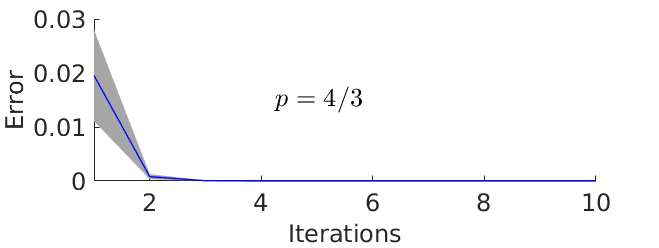}\\
\includegraphics[scale=.25]{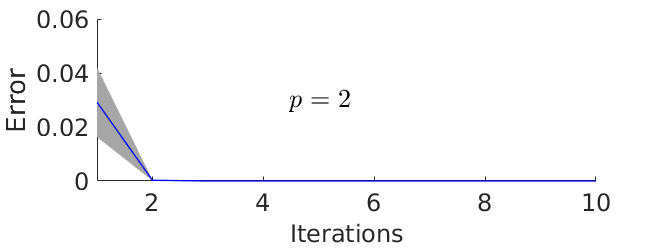}
\includegraphics[scale=.25]{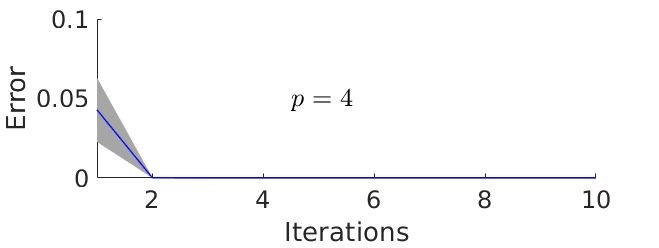}
\includegraphics[scale=.25]{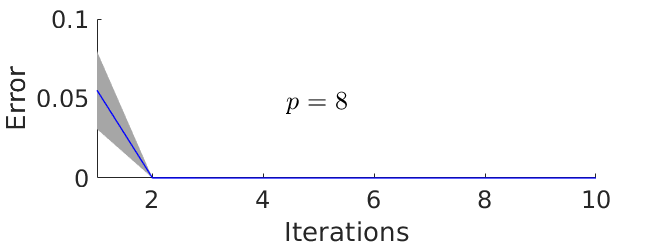}
\includegraphics[scale=.25]{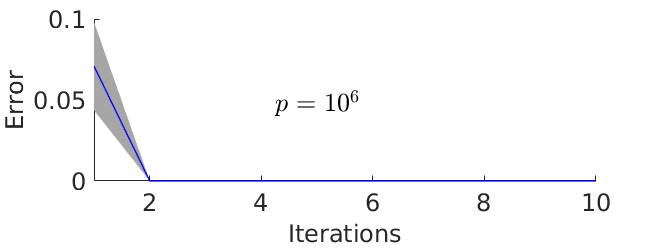}
\caption{Convergence curves for a sample one-class MKL problem for different $p$-norm regularisations (see Section \ref{cb} for further details).}
\label{Concurves}
\end{figure}

In this section, we examine the convergence behaviour of the proposed $\ell_p$-norm one-class MKL approach. For this purpose, we randomly choose a single class from the Oxford Flowers17 \cite{4756141} dataset (a dataset of flower types) and identify it as the target class. Seven distances matrices based on different colour spaces, texture and shape features of flower images are available online \footnote{http://www.robots.ox.ac.uk/$\sim$vgg/data/flowers/17/index.html}. Based on the provided distance matrices, we build seven kernel matrices. Using the proposed approach in Algorithm \ref{generic} ($\delta=10^{-3}n$), we then estimate $\boldsymbol\alpha$ and the optimal kernel weights. As a measure of convergence we record the $l_2$-norm of the change in $\boldsymbol\alpha$. A zero change is indicative of convergence. The experiment is repeated 100 times for each value of $p$ where at each iteration we randomly initialise $\boldsymbol\beta$ to a non-negative vector with a unit $p$-norm. The results are depicted in Fig. \ref{Concurves} where the vertical axis denoted as "Error" represents the $l_2$-norm of the change in $\boldsymbol\alpha$. The solid curve depicts the average while the shaded region represents the standard deviation of the error. From the figure, it can be seen that the convergence is quite fast. More specifically, for smaller values of $p$ (i.e. for $p=32/31$, $16/15$, $8/7$ and $4/3$) convergence is typically attained in as few as 10 iterations. For larger $p$'s, the convergence is even faster where for $p=2$, $4$, $8$ and $10^6$, in most cases, a maximum of 3 iterations suffices for convergence.
\subsection{Effect of different choices for $p$}
\label{peffect}
The goal of the experiments in this section is to illustrate different behaviour of the proposed $\ell_p$ MK-FN approach with regards to different choices for $p$. To this end, two representative values of $p$, namely, $p=2$ and $p=1$ are considered that lead to non-sparse and sparse kernel weights, respectively.
\subsubsection{Synthetic dataset}
In the first part of the experiments in this section, we use a synthetic dataset. We form the target class by sampling 1000 observations from a 2D Gaussian distribution where the mean of the distribution in each dimension is randomly selected between 0 and 1 from a uniform distribution. The covariance of the target distribution is randomly constructed.
\begin{figure}
\centering
\includegraphics[scale=.35]{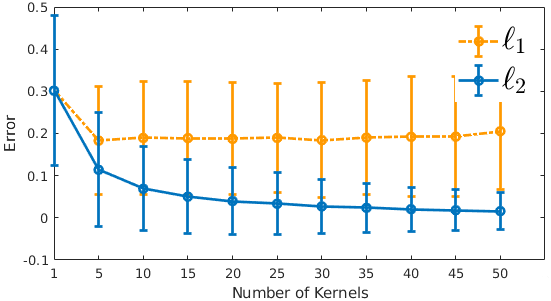}
\caption{Different characteristics of the $\ell_1$- and $\ell_2$-norm regularisation in the proposed method: mean and standard deviation of the classification error as a function of number kernels.}
\label{syn}
\end{figure}
\begin{figure}[t]
\centering
\includegraphics[scale=.4]{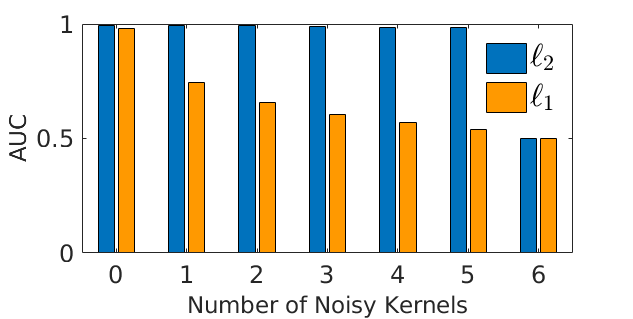}
\caption{Comparison of the performance of a sparse ($\ell_1$-norm) and a non-sparse ($\ell_2$-norm) regularisation scheme in the proposed approach in the presence of noisy kernels on the UCI Multiple Features dataset.}
\label{noisy}
\end{figure}
The test data is formed by sampling 1000 test points, 500 samples of which are drawn from the same distribution as that of the target class to form the positive test data and 500 samples from a different distribution whose mean is drawn from a uniform distribution between 0.5 and 1.5 to form the negative test observations. A validation set consisting of 100 positive and 100 negative samples is randomly generated. The positive validation samples are drawn from the same distribution as that of the target class while the negative validation samples are drawn from the same distribution as that of the negative test data. We repeat the random process above $J$ times, yielding $J$ kernels that can be assumed as kernels each capturing a distinct "view" of the same OCC problem. With this picture in mind, the $\ell_1$ and $\ell_2$ MK-FN approach are used to learn kernel weights. $J$ is varied from 1 to 50 with a step of 5 where at each step the error rates corresponding to the $\ell_1$ and $\ell_2$ MK-FN methods are recorded. The process is above repeated 100 times for each choice of $J$. The mean and standard deviation of the classification errors are plotted in Figure \ref{syn}. From Fig. \ref{syn}, it can be observed that while increasing the number of kernels reduces the error rate for the $\ell_2$ regularisation considerably, it dose not have a big impact on the performance of the $\ell_1$ regularisation. This is understandable as using more kernels is equivalent to bringing in more information and the $\ell_2$-norm may benefit from the additional information by assigning non-sparse weights to different kernels. The $\ell_1$-norm, on the other hand, does not benefit as much from such an increase in the number of kernels due to its over-selective behaviour.
\begin{table*}
\footnotesize
\renewcommand{\arraystretch}{1.2}
\caption{Comparison of the performance of the proposed approach against ten multiple kernel fusion methods on the Delft one-class datasets in terms of AUC (mean$\pm$std $\%$).}
\label{DOC}
\centering
\scriptsize{
\begin{tabular}{lcccccccccc}
\hline
 &\texttt{Vehicle}&\texttt{Diabetes} &\texttt{Sonar}&\texttt{Breast}&\texttt{Imports}&\texttt{Glass}&\texttt{Wine}&\texttt{Liver}&\texttt{Survival}\\
\hline
FN-Average &$90.8\pm3.9$&$57.6\pm4.9$&$57.1\pm3.9$&$97.1\pm1.8$&$66.4\pm8.3$&$83.3\pm6.0$&$99.3\pm0.3$&$62.6\pm5.8$&$66.7\pm7.0$\\
FN-Product	&$90.8\pm2.1$&$57.7\pm6.5$&$60.5\pm4.6$&$92.5\pm4.0$&$73.8\pm9.2$&$82.8\pm6.3$&$99.2\pm1.0$&$61.3\pm5.9$&$62.6\pm7.5$\\
GP-Average &$82.1\pm3.1$&$58.8\pm4.4$&$63.8\pm5.4$&$84.6\pm5.9$&$60.5\pm7.1$&$82.9\pm6.0$&$99.6\pm0.4$&$61.1\pm6.0$&$57.9\pm5.4$\\
GP-Product &$39.8\pm8.0$&$44.7\pm5.7$&$47.0\pm8.6$&$30.8\pm11.4$&$55.4\pm12.0$&$49.8\pm10.9$&$65.0\pm12.7$&$56.5\pm6.7$&$47.2\pm9.1$\\
KPCA-Average &$78.7\pm4.1$&$50.2\pm5.4$&$54.1\pm4.6$&$96.6\pm1.8$&$66.6\pm8.2$&$74.8\pm8.4$&$99.5\pm0.3$&$50.9\pm6.9$&$67.2\pm7.2$\\
KPCA-Product &$82.2\pm3.6$&$49.4\pm6.4$&$57.4\pm4.8$&$86.7\pm4.9$&$75.1\pm9.2$&$74.7\pm8.2$&$99.1\pm0.9$&$50.9\pm6.9$&$64.9\pm7.1$\\
MK-SVDD &$84.9\pm5.2$&$51.5\pm10.2$&$57.0\pm11.2$&$80.9\pm12.7$&$50.0\pm0.0$&$75.4\pm14.3$&$93.5\pm6.1$&$57.6\pm7.2$&$62.3\pm5.9$\\
Slim-MK-SVDD &$89.8\pm4.3$&$58.1\pm5.6$&$63.0\pm7.2$&$96.5\pm1.6$&$72.4\pm8.1$&$82.3\pm7.0$&$99.2\pm0.9$&$58.4\pm7.3$&$65.7\pm6.3$\\
MK-OCSVM &$84.9\pm5.1$&$51.9\pm9.8$&$56.6\pm11.1$&$84.7\pm8.7$&$53.2\pm10.3$&$77.2\pm13.5$&$93.5\pm5.8$&$57.4\pm7.3$&$62.8\pm5.6$\\
Slim-MK-OCSVM &$90.6\pm4.0$&$58.0\pm4.6$&$62.6\pm7.5$&$96.8\pm1.3$&$72.4\pm8.2$&$81.8\pm7.0$&$99.1\pm0.9$&$58.4\pm7.3$&$65.3\pm6.5$\\
$\ell_p$ MK-FN  &$\mathbf{92.8\pm2.8}$&$\mathbf{61.3\pm4.6}$&$\mathbf{66.0\pm5.2}$&$\mathbf{97.6\pm1.3}$&$\mathbf{81.4\pm8.5}$&$\mathbf{84.4\pm5.8}$&$\mathbf{99.7\pm0.3}$&$\mathbf{64.5\pm5.4}$&$\mathbf{70.5\pm4.9}$\\
\hline
\end{tabular}
}
\end{table*}
\subsubsection{UCI Multiple Features dataset}
In this section, we study the impact of different choices for $p$, namely, $p=1$ and $p=2$ on the behaviour of the proposed $\ell_p$ MK-FN approach in the presence of noisy and uninformative kernels on the Multiple Features digit recognition dataset \footnote{Available at https://archive.ics.uci.edu/ml/support/Multiple+Features} from the UCI machine learning repository. The dataset incorporates 6 feature representations for 2000 handwritten
digits. As it is a multi-class dataset, we convert it into a one-class dataset by considering a single class as the normal/target class and all the others as anomalies with respect to the target class. We then repeat this procedure for each class in the dataset, resulting in 10 one-class problems. Using the six representations provided, we build six informative kernels. Additionally, we build six noisy kernels using randomly generated data samples and mix them with the informative ones. We build a multiple kernel fusion system by combining six kernels where the number of noisy kernels within the composite kernel is gradually increased from zero to six. We record the area under the ROC curve (AUC) over all one-class problems (all digits) as a measure of the average performance. The procedure above is repeated 100 times where in each run we randomly select $60\%$ of the positive observations as target training samples, $20\%$ as positive validation data and $20\%$ as positive test observations. For each one-class problem, the negative test and the negative validation samples are selected from all classes but the class of interest. The results corresponding to this experiment are depicted in Fig. \ref{noisy}. From the figure, one may observe that the non-sparse regularisation outperforms the sparse variant for all different numbers of noisy kernels. In particular, while the $\ell_2$ regularisation provides almost stable performance over a wide range of number of noisy kernels, the $\ell_1$ regularisation exhibits relatively inferior performance by locking onto a single, and possibly noisy kernel when the number of uninformative kernels is increased. This is particularly evident when the number of noisy kernels is increased towards 4 or 5 where the $\ell_1$-norm provides a poor performance close to random guess whereas the non-sparse $\ell_2$-norm provides relatively much better resilience against noisy information in the system.
\subsection{Experiment on Benchmark One-Class Datasets}
\label{Delft}
In this section, the results of an evaluation of the proposed method on several benchmark one-class datasets from the repository of the Delft university for one-class classification \cite{delft} are provided. These are the \texttt{Vehicle}, \texttt{Diabetes}, \texttt{Sonar}, \texttt{Breast}, \texttt{Imports}, \texttt{Glass}, \texttt{Wine}, \texttt{Liver} and \texttt{Survival} datasets. Following earlier studies \cite{8259375}, each feature attribute of the provided feature vectors is used to construct a separate kernel. On each dataset, $60\%$ of the data is randomly selected for training while $20\%$ is used for validation and the rest for testing purposes. Each experiment is repeated ten times and the mean and the standard deviations of AUCs' (area under the ROC curves) are reported in Table \ref{DOC}. From Table \ref{DOC}, it may be observed that the FN-average approach (Fisher null-space method operating on the arithmetic mean kernel) provides very competitive performance to the SVM-based multiple kernel learning methods. In particular, the FN-average method typically performs better than the SVM-based multiple kernel learning methods (i.e. MK-SVDD and MK-OCSVM) while providing a close performance to the "slim" variants of these approaches. This is achieved despite using a fixed rule for combining multiple kernels in the FN-average method which emphasises the effectiveness of the Fisher null-space approach for one-class classification. Furthermore, it may be observed that the proposed $\ell_p$ MK-FN approach outperforms other multiple kernel fusion methods on all datasets which illustrates the efficacy of learning kernel combination weights for OCC and also highlights the superiority of the proposed OCC MKL approach compared to other fixed-rule or SVM-based multiple kernel learning methods.
\subsection{Abnormal Image Detection}
\label{Abn}
In abnormal image detection the task is to label images as normal or abnormal. As the form of the abnormality in the images is not known a priori, training is performed on images from the normal/target class. One of the dedicated datasets for visual abnormality detection is the 1001 Abnormal Objects dataset \cite{6618951} that incorporates 1001 abnormal images from 6 object classes that originally appeared in the PASCAL database \cite{everingham2009the}. The groundtruth labels regarding abnormality of an object in this dataset are obtained using human responses collected from the Amazon Mechanical Turk. The six object classes included in the 1001 Abnormal Objects dataset are Boat, Airplane, Chair, Car, Sofa and Motor-bike where the number of abnormal images associated with each object class in the dataset is at least 100. We construct seven kernels for this dataset using deep representations obtained from the pre-trained deep CNN's of Googlenet \cite{7298594}, Resnet50 \cite{He2016DeepRL}, Vgg16 \cite{Simonyan14c}, Alexnet \cite{NIPS2012_c399862d}, Nasnetlarge \cite{8579005}, Mobilenetv2 \cite{8578572} and Densenet201 \cite{8099726}. We follow the protocol associated with this dataset introduced in \cite{6618951} to enable a fair comparison with other methods in the literature. For the comparison with the proposed one-class $\ell_p$ MK-FN approach, we include ten other multiple kernel fusion methods introduced earlier as well as state-of-the-art approaches from the literature including end-to-end deep one-class learning methods. The performances of different methods on this dataset are reported in Table \ref{ImageAbnormality}. From the table, one may observe that the proposed $\ell_p$ MK-FN approach performs better than other multiple kernel methods. The proposed approach achieves an average AUC of $96.2\%$ whereas the second best performing kernel-based method obtains an average AUC of $94.7\%$. It is interesting to note that the second best performing multiple kernel method does not correspond to one of the SVM-based MKL methods but to a simple average fusion of multiple kernels in the Fisher null-space framework. The best reported result in the literature corresponds to the end-to-end one-class deep learning approach of \cite{8721681} with an average AUC of $95.6\%$ which is inferior to the performance of the proposed approach which underlines the utility of an optimal combination of multiple kernels (as in the proposed approach) associated with different deep representations for visual abnormality detection.
\begin{table}
\renewcommand{\arraystretch}{1.2}
\caption{Comparison of different methods for abnormal image detection on the Abnormality-1001 dataset.}
\label{ImageAbnormality}
\centering
\scriptsize{
\begin{tabular}{lc}
\hline
\textbf{Method} & \textbf{AUC (mean$\pm$std $\%$)}\\
\hline
FN-Average &$94.7\pm0.2$\\
FN-Product	&$90.1\pm0.3$\\
GP-Average &$93.5\pm0.2$\\
GP-Product &$84.4\pm0.3$\\
KPCA-Average &$91.6\pm0.2$\\
KPCA-Product &$87.3\pm0.3$\\
MK-SVDD &$92.3\pm0.2$\\
Slim-MK-SVDD &$94.1\pm0.1$\\
MK-OCSVM &$91.4\pm0.3$\\
Slim-MK-OCSVM &$93.9\pm0.1$\\
Graphical Model \cite{6618951} & $87.0\pm n.a.$ \\
Adjusted Graphical Model \cite{6618951}& $91.1\pm n.a.$\\
Autoencoder \cite{1640964} & $67.4\pm1.2$\\
OCNN \cite{DBLP:journals/corr/abs-1802-06360}& $88.5\pm1.4$\\
DOC \cite{8721681} & $95.6\pm3.1$ \\
OC-CNN \cite{8586962}& $84.3\pm n.a.$\\
$\ell_p$ MK-FN &$\mathbf{96.2\pm0.3}$\\
\hline
\end{tabular}}
\end{table}
\subsection{One-Class Novelty Detection}
\label{ImgNovel}
The goal in one-class novelty detection is to gauge the novelty of a new observation using the previously enrolled data items. As the type of novelty in test observations is unknown before the testing stage, training may be performed using one-class classification strategies. Two of the widely used datasets to evaluate the performance of different novelty detection methods are those of the Caltech 256 dataset \cite{256} and the MNIST dataset \cite{lecun-mnisthandwrittendigit-2010}. The Caltech 256 dataset incorporates object images from 256 different classes giving rise to a total of 30607 object images. In order to enable a fair comparison with other methods, we follow two different protocols. The first protocol is the one used in \cite{DBLP:journals/corr/abs-1801-05365} which considers each single class as the target class and assumes samples from all the other 255 object classes as novelties. The experiment is repeated in turn for the first 40 classes in the dataset and the performance is reported in terms of the AUC. The second protocol followed for the Caltech256 dataset is the one employed in \cite{8578454} which considers each one of five randomly chosen categories as the target class and the samples from the "clutter" category as outliers. The second dataset used for novelty detection in this study is that of MNIST \cite{lecun-mnisthandwrittendigit-2010} which is comprised of 60,000 handwritten digits from "0" to "9". For the purpose of this experiment, each digit is regarded as the target class while all other digits serve as novelties. The experiment is repeated in turn for all digits. In this experiment, we report the performances in terms of average AUC over all classes. Multiple kernels are constructed for the images in both datasets using the seven pre-trained deep CNN models similar to the previous experiment. 

Similar to the Abnormal-1001 dataset, for the comparison with the proposed approach, ten other multiple kernel fusion methods as well as state-of-the-art approaches from the literature including end-to-end one-class deep learning methods are considered. The results of this experiment are reported in Table \ref{ImageNovelty} and Table \ref{caltech5} for the first and the second protocol on the Caltech256 dataset, respectively. Table \ref{mnisttable} reports the evaluation results on the MNIST dataset. A number of observations from these experiments are in order. First, in both evaluation settings on the Caltech256 dataset, all the multiple kernel methods outperform other approaches from the literature including deep one-class learning methods of \cite{DBLP:journals/corr/abs-1802-06360,8721681,1640964,8721681} as well as non-deep methods such as R-graph \cite{8099943}. Among the multiple kernel methods, similar to the abnormal image dataset, while the method with the best performance is the proposed $\ell_p$-norm one-class MKL approach, the second best performing method does not correspond to any of the SVM-based MKL approaches but to other fixed-rule multiple kernel methods of the Fisher null-space, GP and the KPCA approaches. The best performing method in both protocols of the Caltech256 dataset is that of the proposed $\ell_p$ MK-FN approach with an average AUC of $99.5\%$ for the 40-class scenario and an average AUC of $99.8$ for the 5-class setting. The best performing deep one-class learning method from the literature for the 40-class setting corresponds to the method in \cite{8721681} with an average AUC of $98.1\%$ while the best performing method from the literature for the 5-category case is that of \cite{8578454} with an average AUC of $92.3\%$. On the MNIST dataset, the proposed $\ell_p$ MK-FN approach obtains the best performance with an average AUC of $98.7\%$ while the best performing methods from the literature correspond to those in \cite{8953440} and \cite{8953541} with an average AUC of $97.5\%$. Similar to the Caltech256 and the Abnormality-1001 datasets, the Fisher null space approach even when using a fixed kernel fusion rule provides very competitive performance to the-state-of-the-art methods which illustrates the efficacy of the Fisher null classification principle for OCC.
\begin{table}
\renewcommand{\arraystretch}{1.2}
\caption{Comparison of different methods for novelty detection on the Caltech 256 dataset using the first protocol.}
\label{ImageNovelty}
\centering
\scriptsize{
\begin{tabular}{lc}
\hline
\textbf{Method} & \textbf{AUC (mean$\pm$std $\%$)}\\
\hline
FN-Average &$99.3\pm0.6$\\
FN-Product	&$99.3\pm0.6$\\
GP-Average &$99.3\pm0.6$\\
GP-Product &$99.3\pm0.6$\\
KPCA-Average &$99.2\pm0.8$\\
KPCA-Product &$99.3\pm0.8$\\
MK-SVDD &$99.1\pm0.7$\\
Slim-MK-SVDD &$99.2\pm0.6$\\
MK-OCSVM &$99.1\pm0.8$\\
Slim-MK-OCSVM &$99.1\pm0.8$\\
OCNN VGG16 \cite{DBLP:journals/corr/abs-1802-06360} &$88.5\pm14.4$\\
OCNN AlexNet \cite{DBLP:journals/corr/abs-1802-06360} &$82.6\pm15.3$\\
OCSVM VGG16 \cite{8721681} &$90.2\pm5.0$\\
Autoencoder \cite{1640964}& $62.3\pm7.2$\\
DOC AlexNet \cite{8721681} &$95.9\pm2.1$\\
DOC VGG16 \cite{8721681} &$98.1\pm2.2$\\
$\ell_p$ MK-FN &$\mathbf{99.5\pm0.5}$\\
\hline
\end{tabular}}
\end{table}

\begin{table}
\renewcommand{\arraystretch}{1.2}
\caption{Comparison of different methods for novelty detection on the Caltech 256 dataset using the second protocol.}
\label{caltech5}
\centering
\scriptsize{
\begin{tabular}{lc}
\hline
\textbf{Method} & \textbf{AUC (mean$\pm$std $\%$)}\\
\hline
FN-Average &$99.6\pm0.4$\\
FN-Product	&$99.5\pm0.3$\\
GP-Average &$99.7\pm0.3$\\
GP-Product &$99.6\pm0.4$\\
KPCA-Average &$99.4\pm0.7$\\
KPCA-Product &$99.4\pm0.6$\\
MK-SVDD &$99.4\pm0.6$\\
Slim-MK-SVDD &$99.4\pm0.4$\\
MK-OCSVM &$99.3\pm0.5$\\
Slim-MK-OCSVM &$99.3\pm0.7$\\
DPCP \cite{7406463} &$67.7\pm n.a.$\\
R-graph \cite{8099943} &$91.3\pm n.a.$\\
The work in \cite{8578454}&$92.3\pm n.a.$\\
$\ell_p$ MK-FN &$\mathbf{99.8\pm0.2}$\\
\hline
\end{tabular}}
\end{table}

\begin{table}
\renewcommand{\arraystretch}{1.2}
\caption{Comparison of different methods for novelty detection on the MNIST dataset.}
\label{mnisttable}
\centering
\scriptsize{
\begin{tabular}{lc}
\hline
\textbf{Method} & \textbf{AUC $\%$}\\
\hline
FN-Average &$97.9$\\
FN-Product	&$93.2$\\
GP-Average &$96.3$\\
GP-Product &$87.7$\\
KPCA-Average &$94.9$\\
KPCA-Product &$90.2$\\
MK-SVDD &$95.7$\\
Slim-MK-SVDD &$97.4$\\
MK-OCSVM &$94.6$\\
Slim-MK-OCSVM &$97.1$\\
The work in \cite{8953541} &$97.5$\\
OCGAN \cite{8953440} &$97.5$\\
the work in \cite{kliger2018novelty}&$97.1$\\
$\ell_p$ MK-FN &$\mathbf{98.7}$\\
\hline
\end{tabular}}
\end{table}
\subsection{Face Presentation Attack Detection}
\label{Anoface}
In face biometrics, an important application of one-class classification is that of presentation attack detection (PAD) where the goal is to determine whether a presentation made to the system sensor corresponds to a genuine (bona fide) biometric trait or is a reproduction of a subject's biometric data (presentation attack) to gain illegitimate access to the system. Typical examples of presentation attacks include printed photo attack and video display attack. As new and previously unseen attacks in the training set may potentially be developed by attackers, the problem is quite a challenging one, emphasising the need for a system with high generalisation capability to detect unseen attacks. One of the promising techniques to deal with unseen attacks is known to be that of one-class classification \cite{7984788}. In this case, bona fide samples are considered as target objects and presentation attacks (PA's) as anomalies. The face presentation attack detection (PAD) problem may be approached in a subject-specific fashion by training a distinct one-class learner for each individual subject enrolled in the dataset \cite{fatemipr}. In this work, the 12 kernels developed in \cite{arashloo2019unseen} are utilised. The kernels are based on the deep representations obtained by applying the pre-trained GoogleNet \cite{7298594}, ResNet50 \cite{He2016DeepRL} and VGG16 \cite{Simonyan14c} networks onto four different facial regions including the whole face as well as different local facial regions that correspond to the main facial features including the joint eyes and the nose region, the nose region and the lower part of the face. We conduct the experiment in an unseen scenario meaning that only positive (i.e. bona fide) samples are utilised for training. The standard metrics for performance reporting in biometric PAD are ISO metrics BSISO-IEC30107-3-2017 \cite{BSISO-IEC30107-3-2017} defined as 1) attack presentation classification error rate (APCER for short) that represents the percentage of attack presentation attempts via the same PA instrument species which are misclassified as bona fide observations; and 2) bona fide presentation classification error rate (BPCER for short) that corresponds to the proportion of bona fide presentation attempts that are misclassified as PA's. The performance of a PAD system in detecting PA samples is reported using the highest APCER among all presentation attack instrument species (PAIS):
\begin{eqnarray}
APCER=\max_{PAIS}APCER_{PAIS}
\end{eqnarray}
The overall performance of the PAD system may be summarised in terms of the Average Classification Error Rate (ACER):
\begin{eqnarray}
ACER = \frac{\max_{PAIS}APCER_{PAIS} + BPCER}{2}
\end{eqnarray}
The face presentation attack dataset used in this study is the OULU-NPU dataset \cite{7961798} which is comprised of 4950 bona fide videos and attack samples from 55 subjects. The samples are partitioned into three subsets of non-overlapping individuals for training (360 bona fide videos and 1440 PA items), development (270 bona fide samples and 1080 PA videos) and testing (360 bona fide samples and 1440 PA videos). Among the four evaluation protocols of the dataset, the most challenging one is the forth protocol which is used in this study to simultaneously evaluate the performance across previously unseen illumination conditions, background scenes, PAIs and input sensors. The evaluation protocol requires to report the average and the standard deviation of the performance over six different mobile devices used to collect the dataset.

The Oulu-NPU dataset lack subject-specific validation data. As such, for the "$\ell_p$ MK-FN" approach, for each subject in the test set, we use the data corresponding to all subjects but the subject of interest as the validation data. For the "Joint $\ell_p$ MK-FN" approach, we use the data associated with the subjects other than the test subjects as the validation set. Apparently, this is not an ideal choice to tune system parameters which is forced by the limitations in terms of data and can potentially lead to a sub-optimal performance.
\begin{figure}[!t]
\centering
\includegraphics[width=3in]{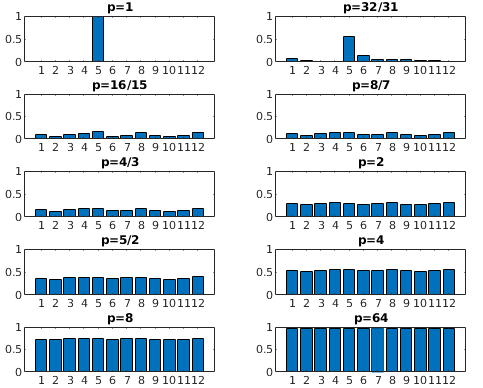}
\caption{Shared kernel weights inferred for the Oulu-NPU dataset for different $p$-norm regularisations.}
\label{oxcal}
\end{figure}
\begin{table}[t]
\renewcommand{\arraystretch}{1.2}
\caption{Comparison of different methods for one-class unseen face presentation attack detection on Protocol IV of the Oulu-NPU dataset.}
\label{ONC}
\centering
\scriptsize{
\begin{tabular}{lcc}
\hline
\textbf{Method} & \textbf{ACER (mean$\pm$std $\%$)}\\
\hline
FN-Average &$5.0\pm3.9$ \\
FN-Product &$4.5\pm5.3$ \\
GP-Average &$6.2\pm4.4$\\
GP-Product & $5.8\pm6.4$\\
KPCA-Average &$5.4\pm3.6$\\
KPCA-Product &$4.5\pm5.3$\\
MK-SVDD &$7.1\pm6.2$\\
Slim-MK-SVDD &$6.2\pm4.4$\\
MK-OCSVM &$7.9\pm6.4$\\
Slim-MK-OCSVM &$6.2\pm4.4$\\
OCA-FAS \cite{QIN2020384} & $4.1\pm2.7$\\
the work in \cite{feng2020learning2} & $3.7\pm2.1$\\
the work in \cite{8737949} & $9.8\pm4.2$\\
SAPLC \cite{9056824} & $9.3\pm4.4$\\
$\ell_p$ MK-FN & $3.3\pm3.4$ \\
Joint $\ell_p$ MK-FN & $\mathbf{3.3\pm3.0}$ \\
\hline
\end{tabular}}
\end{table}
As the one-class problems in this dataset share similarities (i.e. they all correspond to face images) one may infer common kernel weights shared across all one-class problems. The shared kernel weights for the test subjects in this dataset are depicted in Fig. \ref{oxcal} and the ACER's are reported in Table \ref{ONC} where ten multiple kernel methods as well as the best performing end-to-end deep learning-based methods from the literature are included for a comparison. From Table \ref{ONC}, the following observations can be made. First, the fixed-rule one-class Fisher-based and the KPCA-based methods perform better than the SVM-based MKL approaches. Among the SVM-based MK methods, the "slim" variants of the SVDD and the OCSVM methods appear to perform better than their na\"ive versions, as expected. Among the multiple kernel methods, the proposed "$\ell_p$ MK-FN" approach achieves the best performance. Moreover, it outperforms the best reported unseen PAD performance on this dataset corresponding to the method in \cite{feng2020learning2} based on deep end-to-end learning for anomaly detection. The "Joint $\ell_p$ MK-FN" method appears to perform relatively well. In particular, while it archives a similar average error rate as that of the $\ell_p$ MK-FN method, nevertheless, in practical settings, it provides an additional advantage to the $\ell_p$ MK-FN in terms of computational complexity as any new user may directly benefit from the previously learned kernel weights avoiding the requirement for a new subject-specific multiple kernel learning stage.
\subsection{Running Time of Different One-Class MKL Methods}
\label{RunTime}
In this section, a comparison of different one-class MKL methods in terms of their running times is provided. Two sets of experiments are conducted to study the impact of the number of kernels and the number of training samples on training times. Note that as discussed previously, the optimisation problem in Eq. \ref{beta1} is concave, and hence, it may be solved via convex optimisation tools. As an additional method, we solve the minimisation subproblem in Eq. \ref{beta1} using CVX, a package for specifying and solving convex programmes \cite{cvx} and refer to it as $\ell_p$ MK-FN-CVX in the comparisons. Moreover, we examine a gradient ascent-based approach for optimisation which is referred to as $\ell_p$ MK-FN-Grad (please consult Section V of the supplementary material for a derivation of the gradient of the cost function). Note that as we set the stopping condition for the $\ell_p$ MK-FN, $\ell_p$ MK-FN-CVX and the $\ell_p$ MK-FN-Grad methods similarly, their results exactly match, and hence, their running times are comparable. We generate the data for this experiment similar to Section \ref{peffect} where $p$ is set to 2. The proposed method is implemented as un-optimised Matlab codes. The implementation of the SVM-based MKL methods are due to \cite{loosli:hal-01593595}. Each experiment is repeated 100 times on a 64-bit 4.00GHz Intel Core-i7 machine with 32GB memory and we report the average training times over 10 trials. The results of this experiment are reported in Table \ref{T1} and Table \ref{T2}. In order to better compare the running time of different one-class MKL algorithms in the aforementioned tables we have excluded the time required to compute the kernel matrix which is common to all algorithms and has a time complexity of $\mathcal{O}(n^2)$ on a serial machine. Nevertheless, kernel matrix computation may benefit from a parallel processing hardware such as a GPU to achieve large speed-up gains.

\begin{table}
\renewcommand{\arraystretch}{1.2}
\caption{Average training times of different one-class multiple kernel learning methods (in milliseconds) for different number of kernels ($J$). (the number of training samples is 100, their dimensionality is 100.)}
\label{T1}
\centering
\begin{tabular}{lccccc}
\hline
\textbf{Number of kernels} & $5$ & $15$ & $50$& $100$& $500$\\
\hline
$\ell_p$ MK-FN &1.8&4.2&13.9&4.2&33.4\\
$\ell_p$ MK-FN-Grad &35.4&410.5&420.5&4893.2&115881.3\\
$\ell_p$ MK-FN-CVX &334.5&321.4&346.4&516.0&1177.9\\
Slim-MK-SVDD &34.1&65.3&296.1&589.1&1696.0\\
MK-SVDD &78.5&173.7&325.6&1228.9&4094.6\\
Slim-MK-OCSVM &41.9&93.7&221.4&982.1&685.8\\
MK-OCSVM &49.2&85.4&324.7&1293.4&2608.9\\
\hline
\end{tabular}
\end{table}
\begin{table*}
\renewcommand{\arraystretch}{1.2}
\caption{Average training times of different one-class multiple kernel learning methods (in milliseconds) for different number of training samples ($n$). (the dimensionality of training samples is 100 and the number of kernels is set to 10.)}
\label{T2}
\centering
\begin{tabular}{lcccccc}
\hline
\textbf{No. of training samples} & $50$& $100$ & $1000$ &$2000$& $5000$&$10000$\\
\hline
$\ell_p$ MK-FN &0.8&2.1&719.0&758.8&6820.8&21315.7\\
$\ell_p$ MK-FN-Grad &18.0&58.3&20713.0&46727.1&135635.4&323040.8\\
$\ell_p$ MK-FN-CVX &330.0&529.2&1066.8&1668.7&10387.3&37022.8\\
Slim-MK-SVDD &453.9&605.7&30392.2&107384.8&85023.3&497251.3\\
MK-SVDD &32.9&63.5&2172.2&4768.4&26094.2&108472.1\\
Slim-MK-OCSVM &199.4&404.0&39973.4&85429.2&401301.7&1753212.8\\
MK-OCSVM &22.3&46.7&4254.1&12757.6&55474.9&163852.4\\
\hline
\end{tabular}
\end{table*}
From the tables the following observations are in order. The proposed $\ell_p$ MK-FN approach when compared to the $\ell_p$ MK-FN-CVX and the MK-FN-Grad is faster by multiple orders of magnitude. When compared to the SVM-based MKL methods, the proposed $\ell_p$ MK-FN method also comes out as the most efficient method in the majority of the experiments. In summary, the proposed $\ell_p$ MK-FN method outperforms other MKL methods in different scenarios considered above.
The work in \cite{8099922} presents an efficient one-class Fisher null classifier via an incremental approach. However, due to the absence of a Tikhonov regularisation on the solution, it may not be directly plugged into the proposed approach since the convergence of the proposed MKL algorithm (discussed in Section III of the supplementary material) heavily relies on the Tikhonov regularisation. An investigation on the modifications required to be applied on the algorithm presented in \cite{8099922} so that it could be employed within the proposed MKL framework may be considered as a future direction of investigation.

\subsubsection{Scalability}
As observed above, among other one-class multiple kernel learning algorithms, the proposed approach is computationally more efficient. In addition, even compared to the state-of-the-art convex optimisation tools, the proposed approach is faster as the proposed optimisation algorithm is better tailored to the problem at hand. Furthermore, the proposed approach is also faster than a gradient-based optimisation approach as observed in tables \ref{T1} and \ref{T2}.

The computational complexity of the proposed method is dominated by the computation of vector $\mathbf{u}$ requiring matrix-vector multiplications (step 4 of Algorithm \ref{generic}), the time complexity of which is $\mathcal{O}(Jn^2)$ where $J$ is the number of kernels and $n$ denotes the number of training samples. Note that a na\"ive computation of the inverse matrix in step 6 of Algorithm \ref{generic} would require $\mathcal{O}(n^3)$ time. However, the computational complexity of matrix inversion may be reduced to $\mathcal{O}(n^2)$ using an incremental Cholesky decomposition based on the Sherman's march algorithm \cite{matrixdecom,6905848}.

An appealing characteristic of the proposed method is that all the operations involved are parallelizable. In particular, the computation of vector $\mathbf{u}$ (step 4 of Algorithm \ref{generic}) that involves matrix-vector multiplications can be readily ported onto parallel processing units such as GPU's to obtain massive speed-up gains. Similarly, there has been a large volume of work on parallel implementations of the matrix inversion operation where it is shown that significant improvements in the running time may be obtained \cite{https://doi.org/10.1002/cpe.2933,7418328}.

The parallelizability attribute discussed above also applies to the \textit{Joint} $\ell_p$-norm MKL approach. Moreover, the joint approach provides a particularly appealing additional property which may be deployed for reduced time complexity: under the assumption that the one-class MKL problems share similarities in terms of feature space representations, one may learn the \textit{shared} kernel combination weights for a number of OCC problems offline and directly apply the inferred weights to a new MKL OCC problem, circumventing the computational overhead corresponding to the new OCC MKL problem. As an instance, in the face PAD problem, using a number of subjects previously enrolled into the system, the optimal kernel combination weights may be inferred using the \textit{Joint} $\ell_p$-norm MKL algorithm. A new subject to be registered to the system may then benefit from the common kernel combination weights previously learned, completely avoiding any new one-class multiple kernel learning procedure.
\section{Conclusion}
\label{C}
We addressed the MKL problem for one-class classification. For this purpose, based on the one-class Fisher null approach, an $\ell_p$-norm one-class MKL method was presented. The associated optimisation problem was posed as a saddle point Lagrangian optimisation problem which was then solved via a new fixed-point iteration algorithm. The proposed approach was also extended to jointly learn multiple related one-class problems by constraining them to share common kernel weights. We theoretically studied the extreme cases of the proposed approach and illustrated that the $\ell_1$-norm only selects a single kernel from among a set of candidate kernels while the $\ell_{\infty}$ yields uniform kernel weights. 

The evaluation of the proposed method on a range of datasets from different application domains illustrated the merits of the proposed method against other alternatives. In particular, it was shown that a fixed-norm solution may be outperformed by learning the intrinsic sparsity of the problem at hand as considered in the proposed approach. The proposed $\ell_p$ MK-FN approach was also demonstrated to perform better than the one-class SVM-based MKL methods as well as leading one-class end-to-end deep learning approaches from the literature. As future directions of investigation one may consider a mixed $(r,p)$-norm regularisation scheme on kernel weights (similar in spirit to the work in \cite{8259375}) to provide further flexibility to the one-class MKL model.


%

\ifCLASSOPTIONcaptionsoff
  \newpage
\fi



%
\bibliographystyle{IEEEtran}
\bibliography{IEEEexample}

%






\end{document}


%
\title{Supplementary Material}
%
%
%

\author{Shervin~Rahimzadeh~Arashloo\IEEEmembership{}        
\thanks{}
\thanks{}
\thanks{}}

%
%

\markboth{Journal of \LaTeX\ Class Files,~Vol.~?, No.~?, ?~2020}%
{Shell \MakeLowercase{\textit{et al.}}: Bare Demo of IEEEtran.cls for IEEE Journals}
%



\maketitle



%
\IEEEpeerreviewmaketitle

\section{}
\label{app1}
In this section, we show that the projection $\phi(x_i)^\top\mathbf{w}$ defined in terms of the optimal solution of Eq. 6 yields a kernel Fisher null-space projection for one-class classification. The problem in Eq. 6 may be written as
\begin{eqnarray}
\nonumber \min_\mathbf{w} \frac{1}{n}\sum_{i=1}^n(1-\boldsymbol\phi(\mathbf{x}_i)^\top\mathbf{w})^2=\min_\mathbf{w}\frac{1}{n}\norm{\mathbf{1}-\boldsymbol\Phi(\mathbf{X})^\top\mathbf{w}}_2^2
\end{eqnarray}
\noindent where $\mathbf{1}$ is an $n$-vector of $1$'s while $\boldsymbol\Phi(\mathbf{X})$ denotes a matrix collection of positive training samples in the RKHS space. The optimal $\mathbf{w}$ may be found by taking the derivative of the cost function w.r.t $\mathbf{w}$ and setting it to zero, yielding
\begin{eqnarray}
\nonumber \mathbf{w} = (\boldsymbol\Phi(\mathbf{X})\boldsymbol\Phi(\mathbf{X})^\top)^{-1}\boldsymbol\Phi(\mathbf{X})\mathbf{1}
\end{eqnarray}
Having found $\mathbf{w}$, the responses of the training observations (denoted as $\mathbf{y}$) may be determined as
\begin{eqnarray}
\nonumber \mathbf{y}=\boldsymbol\Phi(\mathbf{X})^\top\mathbf{w}=\boldsymbol\Phi(\mathbf{X})^\top(\boldsymbol\Phi(\mathbf{X})\boldsymbol\Phi(\mathbf{X})^\top)^{-1}\boldsymbol\Phi(\mathbf{X})\mathbf{1}=\mathbf{B}\mathbf{1}
\end{eqnarray}
where $\mathbf{B}=\boldsymbol\Phi(\mathbf{X})^\top(\boldsymbol\Phi(\mathbf{X})\boldsymbol\Phi(\mathbf{X})^\top)^{-1}\boldsymbol\Phi(\mathbf{X})$. It can be easily shown that $\mathbf{B}$ is an identity matrix. For this purpose, let us compute $\boldsymbol\Phi(\mathbf{X})\mathbf{B}$:
\begin{eqnarray}
\nonumber \boldsymbol\Phi(\mathbf{X})\mathbf{B}=[\boldsymbol\Phi(\mathbf{X})\boldsymbol\Phi(\mathbf{X})^\top][\boldsymbol\Phi(\mathbf{X})\boldsymbol\Phi(\mathbf{X})^\top]^{-1}\boldsymbol\Phi(\mathbf{X})=\boldsymbol\Phi(\mathbf{X})
\end{eqnarray}
Since $\boldsymbol\Phi(\mathbf{X})\mathbf{B}=\boldsymbol\Phi(\mathbf{X})$, hence $\boldsymbol\Phi(\mathbf{X})(\mathbf{B}-\mathbf{I})=\mathbf{0}$ and consequently $\boldsymbol\Phi(\mathbf{X})^\top\boldsymbol\Phi(\mathbf{X})(\mathbf{B}-\mathbf{I})=\mathbf{0}$. However, $\boldsymbol\Phi(\mathbf{X})^\top\boldsymbol\Phi(\mathbf{X})$ is the kernel matrix by definition and as a result $\mathbf{K}(\mathbf{B}-\mathbf{I})=\mathbf{0}$. Using an RBF kernel  and assuming that not all training samples coincide, the kernel matrix $\mathbf{K}$ shall be positive definite and invertible, and hence, $(\mathbf{B}-\mathbf{I})=\mathbf{K}^{-1}\mathbf{0}=\mathbf{0}$ and as a result  $\mathbf{B}$ is an identity matrix. Consequently, $\mathbf{y}=\mathbf{B}\mathbf{1}=\mathbf{1}$. That is, the responses $\mathbf{y}$ derived using the optimal solution of Eq. 6 are all equal to $1$. In the next section, it is shown that obtaining equal responses for all target training samples is equivalent to the Fisher null-space method for OCC.
\section{}
In a Fisher classifier, the criterion function is
\begin{eqnarray}
\nonumber \mathbf{J}=\frac{\mathbf{\varphi^\top S_b \varphi}}{\mathbf{\varphi^\top S_w \varphi}}
\end{eqnarray}
For the two-category case, the Fisher criterion may be equivalently written as the ratio of the between-class variance to the within-class variance \cite{Alpaydin14}:
\begin{eqnarray}
\nonumber \mathbf{J}=\frac{(M_1-M_2)^\top(M_1-M_2)}{S_1^2+S_2^2}
\end{eqnarray}
where $M_1$ and $M_2$ denote the means of the two classes while $S_1$ and $S_2$ represent the within-class variances. The analysis here employs the second equivalent representation of the Fisher criterion. As $M_1$ denotes the mean of the positive samples after projection (i.e. the mean  of $\mathbf{y}$), we have $M_1=1$. As a result, the within-class variance of the projected positive samples is
\begin{eqnarray}
\nonumber  S_1^2=\sum_{i=1}^n(y_i-M_1)^\top(y_i-M_1)=\sum_{i=1}^n(1-1)^\top(1-1)=0
\end{eqnarray}
where $y_i$ denotes the projection of sample $x_i$ onto an optimal feature subspace. As for the negative class, it is assumed that only a single hypothetical sample exists at the origin, thus $M_2=0$ and $S_2=0$.\\
Regarding the numerator of $\mathbf{J}$ one has
\begin{eqnarray}
\nonumber (M_1-M_2)^{\top}(M_1-M_2)=(1-0)^\top(1-0)=1
\end{eqnarray}
Consequently, the between-class scatter is positive (i.e. $(M_1-M_2)^{\top}(M_1-M_2)=1$) while the total within-class scatter is zero (i.e. $S_1^2+S_2^2=0$) and hence the projection $\boldsymbol\phi(\mathbf{x})^\top\mathbf{w}$ that minimises Eq. 6 yields a kernel Fisher null-space projection for OCC.
\section{}
In this section, we analyse the convergence properties of the proposed $\ell_p$-norm multiple kernel Fisher null-space approach summarised as Algorithm 1. Let us first consider the following definition and the theorem that follows.
\begin{defi}
If there exists a real number $L\geq0$ such that $\forall$ $x$, $x^\prime$ $\in$ $X$, function $f:X\rightarrow Y$  satisfies
\begin{eqnarray}
\nonumber \norm{f(x)-f(x^\prime)}_2\leq L\norm{x-x^\prime}_2
\end{eqnarray}
\end{defi}
\noindent then $f(.)$ is \textit{Lipschitz continuous} over $X$ and $L$ is a \textit{Lipschitz constant} of $f(.)$ \cite{lips}.
The following theorem from \cite{NM} provides sufficient conditions for the convergence of a fixed-point iteration approach.
\begin{thm}
If $f(.)$ is Lipschitz continuous with a Lipschitz constant smaller than one (i.e. $L<1$) then:\\
i) the sequence $\{x_m,m=1,2,\ldots\}$ generated by the iteration $x_{m+1}=f(x_{m})$ will converge towards a fixed-point for any initial guess $x_0$ of the solution;\\
ii) the fixed point is unique.
\label{t1}
\end{thm}
\textbf{Proof}\\
i) As $f(.)$ is Lipschitz continuous and $L<1$, then for $x_m,m=0,1,2,\ldots$, we have
\begin{align}
\nonumber &\norm{x_2-x_1}_2=\norm{f(x_1)-f(x_0)}_2\leq L\norm{x_1-x_0}_2\\
\nonumber &\norm{x_3-x_2}_2=\norm{f(x_2)-f(x_1)}_2\leq L\norm{x_2-x_1}_2\\
\nonumber &\vdots\\
\nonumber &\norm{x_{m+1}-x_{m}}_2=\\ \nonumber&\hspace{1.5cm}\norm{f(x_{m})-f(x_{m-1})}_2\leq L\norm{x_{m}-x_{m-1}}_2
\end{align}
Combining the inequalities above yields
\begin{eqnarray}
\nonumber \norm{x_{m+1}-x_{m}}_2\leq L^{m} \norm{x_1-x_0}_2
\end{eqnarray}
As $L<1$ then $\lim_{m\rightarrow\infty}L^{m}=0$, and consequently, $\{x_m,m=1,2,\ldots\}$ converges to a point.\\
ii) Let us suppose that there are two distinct fixed-points $x^\star$ and $x^\bullet$. Since the function is Lipschitz continuous with a Lipschitz constant smaller than 1, we have:
\begin{eqnarray}
\nonumber \norm{x^\star-x^\bullet}_2=&\\
\nonumber \norm{f(x^\star)-f(x^\bullet)}_2&\leq L\norm{x^\star-x^\bullet}_2<\norm{x^\star-x^\bullet}_2
\end{eqnarray}
Thus, we have arrived at the contradiction $\norm{x^\star-x^\bullet}_2<\norm{x^\star-x^\bullet}_2$. As a result, there cannot be two distinct fixed points and the fixed point is unique.

We now provide the convergence requirements of the proposed method in the following proposition followed by its proof.
\begin{prop}
For a sufficiently large regularisation parameter $\delta$, function $f(\boldsymbol\alpha)=\Big(\delta \mathbf{I}+\sum_{j=1}^J\beta_j\mathbf{K}_j\Big)^{-1}\mathbf{1}$ would be Lipschitz continuous with a Lipschitz constant smaller than $1$, and hence, according to Theorem \ref{t1}, the proposed approach converges to a unique fixed point for any initial guess of $\boldsymbol\alpha$.
\end{prop}

\noindent \textbf{Proof}\\For the proof, it suffices to show that for a sufficiently large $\delta$, it holds that
\begin{eqnarray}
\nonumber \frac{\norm{f(\boldsymbol\alpha^{m+1})-f(\boldsymbol\alpha^{m})}_2} {\norm{\boldsymbol\alpha^{m+1}-\boldsymbol\alpha^{m}}_2}<1
\label{cc}
\end{eqnarray}
\noindent where $\boldsymbol\alpha^{m}$ and $\boldsymbol\alpha^{m+1}$ correspond to two consecutive solutions produced by the proposed algorithm at iterations $m$ and $m+1$. Let us assume $\norm{\boldsymbol\alpha^{m+1}-\boldsymbol\alpha^{m}}_2=\omega$. If $\omega=0$, it follows that the sequence $\boldsymbol\alpha^m$ is converged. Hence, we consider the case when $\omega$ is strictly positive, i.e. $\omega>0$. Considering the numerator of the quotient in Eq. \ref{cc}, we have
\begin{align}
\nonumber &\norm{f(\boldsymbol\alpha^{m+1})-f(\boldsymbol\alpha^{m})}_2\leq\norm{f(\boldsymbol\alpha^{m+1})}_2+\norm{f(\boldsymbol\alpha^{m})}_2\\
\nonumber = &\norm{\Big(\delta \mathbf{I}+\sum_{j=1}^J\beta^{m+1}_{j}\mathbf{K}_j\Big)^{-1}\mathbf{1}}_2+\norm{\Big(\delta \mathbf{I}+\sum_{j=1}^J\beta^{m}_j\mathbf{K}_j\Big)^{-1}\mathbf{1}}_2\\
\nonumber\leq &\sqrt{n}\norm{\Big(\delta \mathbf{I}+\sum_{j=1}^J\beta^{m+1}_{j}\mathbf{K}_j\Big)^{-1}}_2+\sqrt{n}\norm{\Big(\delta \mathbf{I}+\sum_{j=1}^J\beta^{m}_{j}\mathbf{K}_j\Big)^{-1}}_2\\
\nonumber
\end{align}
\noindent where the first inequality is based on the triangle inequality and the last inequality follows from the fact that the norm of a matrix-vector product is upper bounded by the product of the norms of the matrix and the vector \cite{matrixdecom}. Moreover, it is known that the $l_2$-norm of a square matrix is equal to the largest eigenvalue of the matrix and the $l_2$-norm of an inverted matrix is $1/\sigma$ where $\sigma$ is the smallest eigenvalue of the original matrix \cite{matrixdecom}. As a result
\begin{eqnarray}
\nonumber \norm{f(\boldsymbol\alpha^{m+1})-f(\boldsymbol\alpha^{m})}_2\leq\sqrt{n}(1/\sigma_1+1/\sigma_2)
\end{eqnarray}
\noindent where $\sigma_1$ and $\sigma_2$ correspond to the minimum eigenvalues of $(\delta \mathbf{I}+\sum_{j=1}^J\beta^{m+1}_{j}\mathbf{K}_j\Big)$ and $(\delta \mathbf{I}+\sum_{j=1}^J\beta^{m}{j}\mathbf{K}_j\Big)$, respectively. Adding a constant to the main diagonal of a matrix shifts the eigenvalues by the same constant. For the proof, let us assume matrix $\mathbf{K}$ has an eigenvector $\mathbf{z}$ with the corresponding eigenvalue $\eta$, hence $\mathbf{K}\mathbf{z}=\eta\mathbf{z}$. If one adds a constant $c$ to the main diagonal elements of $\mathbf{K}$, it yields $\mathbf{K}^\prime=\mathbf{K}+c\mathbf{I}$ where $\mathbf{I}$ denotes an identity matrix. Let us calculate $\mathbf{K}^\prime\mathbf{z}$:
\begin{eqnarray}
\nonumber \mathbf{K}^\prime\mathbf{z}=(\mathbf{K}+c\mathbf{I})\mathbf{z}=\mathbf{K}\mathbf{z}+c\mathbf{I}\mathbf{z}=\eta\mathbf{z}+c\mathbf{z}=(\eta+c)\mathbf{z}
\end{eqnarray}
As a result, $\mathbf{z}$ is an eigenvector of $\mathbf{K}^\prime$ with the eigenvalue of $\eta+c$.

Consequently, if the minimum eigenvalues of the combined kernel matrices $\sum_{j=1}^J\beta^{m+1}_{j}\mathbf{K}_j$ and $\sum_{j=1}^J\beta^{m}_{j}\mathbf{K}_j$ are denoted as $\sigma^{\prime}_1$ and $\sigma_2^{\prime}$, respectively, we have $\sigma_1=\sigma^{\prime}_1+\delta$ and $\sigma_2=\sigma^{\prime}_2+\delta$. In the worst case scenario where the combined kernel matrices' minimum eigenvalues are zero, one has $\sigma_1=\delta$ and $\sigma_2=\delta$ and as a result
\begin{eqnarray}
\nonumber \norm{f(\boldsymbol\alpha^{m+1})-f(\boldsymbol\alpha^{m})}_2\leq2\sqrt{n}/\delta
\label{upp1}
\end{eqnarray}
Plugging the upper bound given in Eq. \ref{upp1} into Eq. \ref{cc} gives the sufficient condition for the convergence of the proposed algorithm to a unique fixed point:
\begin{eqnarray}
\nonumber \frac{2\sqrt{n}/\delta}{w}<1
\end{eqnarray}
or equivalently
\begin{eqnarray}
\nonumber \delta>\frac{2\sqrt{n}}{w}
\label{dbound}
\end{eqnarray}
For a sufficiently large $\delta$, the above inequality holds true and as a result $f(\boldsymbol\alpha)$ would be Lipschitz continuous with a Lipschitz constant smaller than one. Consequently, according to Theorem \ref{t1}, the proposed algorithm would converge to a unique fixed point regardless of any particular initialisation.$\blacksquare$

In practice, using an RBF kernel function and assuming that no training samples coincide, the combined kernel matrix will be strictly positive definite (i.e. $\sigma^{\prime}_1>0$ and $\sigma^{\prime}_2>0$). In this case, the lower bound for $\delta$ that will ensure convergence will be smaller than that of Eq. \ref{dbound} since $\sigma_1$ and $\sigma_2$ will be larger than $\delta$ by $\sigma^{\prime}_1$ and $\sigma^{\prime}_2$, respectively.

\section{}
\label{app3}
In this section, it is shown that $\boldsymbol\beta^\top\mathbf{u}=\norm{\mathbf{u}}_{p/(p-1)}$.
\\
From Eq. 24, we have $\boldsymbol\beta = \frac{\mathbf{u}^{1/(p-1)}}{\norm{\mathbf{u}^{1/(p-1)}}_p}$ and hence:
\begin{eqnarray}
\nonumber\boldsymbol\beta^\top\mathbf{u}&=&\frac{[\mathbf{u}^{1/(p-1)}]^\top\mathbf{u}}{\norm{\mathbf{u}^{1/(p-1)}}_p}=\frac{\sum_{j=1}^Ju_j^{1/(p-1)}u_j}{\Big[\sum_{j=1}^J u_j^{p/(p-1)}\Big]^{1/p}}\\
&=&\frac{\Big\{\big\{\sum_{j=1}^Ju_j^{p/(p-1)}\big\}^{(p-1)/p}\Big\}^{p/(p-1)}}{\Big\{\big[\sum_{j\nonumber=1}^Ju_j^{p/(p-1)}\big]^{(p-1)/p}\Big\}^{p/(p-1)\times1/p}}\\
\nonumber&=&\frac{\norm{\mathbf{u}}_{p/(p-1)}^{p/(p-1)}}{\norm{\mathbf{u}}_{p/(p-1)}^{1/(p-1)}}=\norm{\mathbf{u}}_{p/(p-1)}
\end{eqnarray}
\\
By plugging the above equality into the cost function, the objective function shall be
\begin{eqnarray}
\nonumber\max_{\boldsymbol{\alpha}}-\delta \boldsymbol{\alpha}^\top \boldsymbol{\alpha}+2\boldsymbol{\alpha}^\top\mathbf{1}-\norm{\mathbf{u}}_{p/(p-1)}
\label{co}
\end{eqnarray}
\section{}
The gradient of the cost function in Eq. 26 may be derived as
\begin{eqnarray}
\nonumber&&\frac{\partial}{\partial\boldsymbol\alpha}
\Big[ -\delta \boldsymbol{\alpha}^\top\boldsymbol{\alpha}+2\boldsymbol{\alpha}^\top\mathbf{1}-\norm{\mathbf{u}}_{p/(p-1)}\Big]\\
\nonumber &=&-2\delta\boldsymbol\alpha+2\times\mathbf{1}-\frac{\partial}{\partial\boldsymbol\alpha}\Big[\sum_{j=1}^Ju_j^{p/(p-1)}\Big]^{(p-1)/p}\\
\nonumber &=&-2\delta\boldsymbol\alpha+2\times\mathbf{1}-(p-1)/p\times\Big[\sum_{j=1}^J\frac{\partial}{\partial\boldsymbol\alpha}{(\boldsymbol\alpha^\top\mathbf{K}_j\boldsymbol\alpha)}^{p/(p-1)}\Big]\\\nonumber&\times&\Big[\sum_{j=1}^Ju_j^{p/(p-1)}\Big]^{-1/p}\\
\nonumber &=&2\Big[-\delta\boldsymbol\alpha+\mathbf{1}-\norm{\mathbf{u}}_{p/(p-1)}^{-1/(p-1)}\Big(\sum_{j=1}^Ju_j^{1/(p-1)}\mathbf{K}_j\Big)\boldsymbol\alpha\Big]
\end{eqnarray}
\\
A gradient ascent approach to perform optimisation shall be
$\boldsymbol\alpha_{t+1}=\boldsymbol\alpha_{t}+\kappa\frac{\partial}{\partial\boldsymbol\alpha}Cost$ where $\kappa$ stands for the step size.
\\
\\


\bibliographystyle{IEEEtran}
\bibliography{ex}